\documentclass[10pt,journal,compsoc]{IEEEtran}

\ifCLASSOPTIONcompsoc
  \usepackage[nocompress]{cite}
\else
  \usepackage{cite}
\fi
\ifCLASSINFOpdf
\else
\fi

\usepackage{graphicx}
\usepackage{dcolumn}
\usepackage{bm}
\usepackage{color}
\usepackage{xcolor}
\usepackage{tabularx}
\usepackage{makecell}
\usepackage{longtable}
\usepackage{bibunits}
\usepackage{times}
\usepackage{amsmath}
\usepackage{cite}
\usepackage{multirow}
\usepackage{support-caption}
\usepackage{subcaption}
\usepackage{caption}
\usepackage{mwe}
\usepackage{bbm}
\usepackage{ctable}
\usepackage{svg}
\usepackage{dutchcal}
\usepackage{verbatim}
\usepackage{lineno}
\begin{document}
\renewcommand{\arraystretch}{1.1}

%
\title{\bfseries\Large A Machine Learning Paradigm for Studying Pictorial Realism:\\ How Accurate Are Constable's Clouds?}
%
%
%
%

\author{Zhuomin~Zhang, Elizabeth~C.~Mansfield, Jia~Li,~\IEEEmembership{Fellow,~IEEE,} John~Russell, George~S.~Young, Catherine~Adams, Kevin~A.~Bowley, and~James~Z.~Wang,~\IEEEmembership{Senior Member,~IEEE}\thanks{The authors are with The Pennsylvania State University, University Park, PA 16802. Correspondence should be addressed to J. Z. Wang (jwang@psu.edu).}}

%
%

\newcommand{\red}[1]{\textcolor{red}{#1}}
\definecolor{shadecolor}{rgb}{0.6,0.6,0.6}
\renewcommand\fbox{\fcolorbox{shadecolor}{white}}

\renewcommand{\textcolor}[2]{#2}

\markboth{}%
{Zhang \MakeLowercase{\textit{{\it et al.}}}: A Machine Learning Paradigm for Studying Pictorial Realism: How Accurate Are Constable's Clouds?}
%



\IEEEtitleabstractindextext{%
\begin{abstract}
The British landscape painter John Constable is considered foundational for the Realist movement in 19\textsuperscript{th}-century European painting. Constable's painted skies, in particular, were seen as remarkably accurate by his contemporaries, an impression shared by many viewers today. Yet, assessing the accuracy of realist paintings like Constable's is subjective or intuitive, even for professional art historians, making it difficult to say with certainty what set Constable's skies apart from those of his contemporaries. Our goal is to contribute to a more objective understanding of Constable's realism. We propose a new machine-learning-based paradigm for studying pictorial realism in an explainable way. Our framework assesses realism by measuring the similarity between clouds painted by artists noted for their skies, like Constable, and photographs of clouds. The experimental results of cloud classification show that Constable approximates more consistently than his contemporaries the formal features of actual clouds in his paintings. The study, as a novel interdisciplinary approach that combines computer vision and machine learning, meteorology, and art history, is a springboard for broader and deeper analyses of pictorial realism.

\end{abstract}

}

\maketitle

\IEEEdisplaynontitleabstractindextext

%
\IEEEpeerreviewmaketitle

\IEEEraisesectionheading{\section{Introduction}\label{sec:introduction}}
%
%
%
%
\IEEEPARstart{I}{N} this paper, we propose a new machine learning paradigm for studying the European art style known as realism. The specific case study we report here is the work of John Constable (1776-1837) whose landscape paintings are considered foundational for the Realist movement. Constable was especially renowned for his skies. Although there is general agreement that Constable's sky paintings are persuasive in their realism, the precise basis for his realism continues to be debated. The feasibility of quantitative analysis for studying pictorial realism, as exemplified here, demonstrates that computational approaches may augment traditional approaches to art-historical research.



\begin{figure}[ht!]
\centering
\begin{subfigure}{.48\linewidth}
  \centering
  \includegraphics[width=\linewidth]{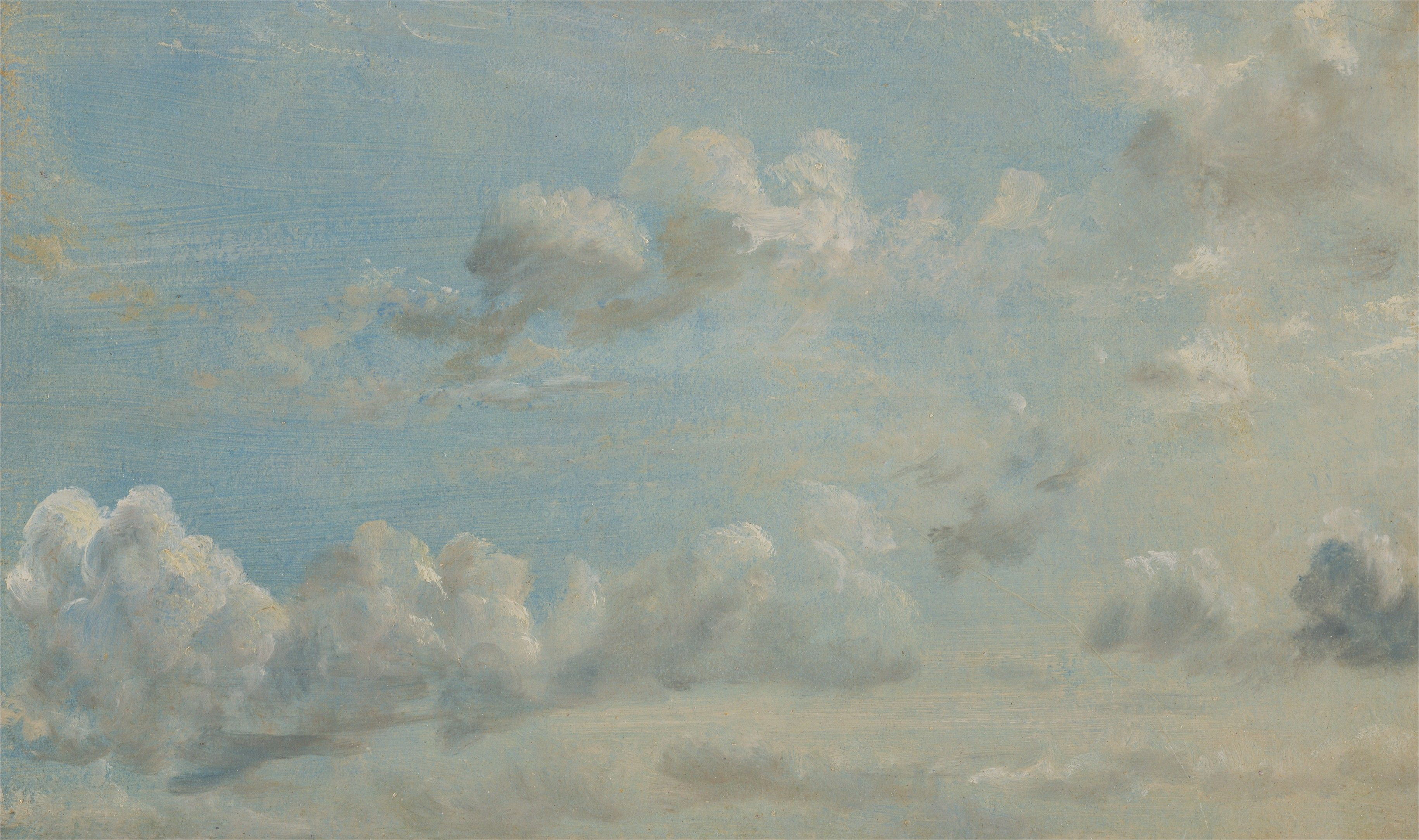}
  \label{fig:sub1}
\end{subfigure}\hspace{0.1in}
\begin{subfigure}{.465\linewidth}
  \centering
  \includegraphics[width=\linewidth]{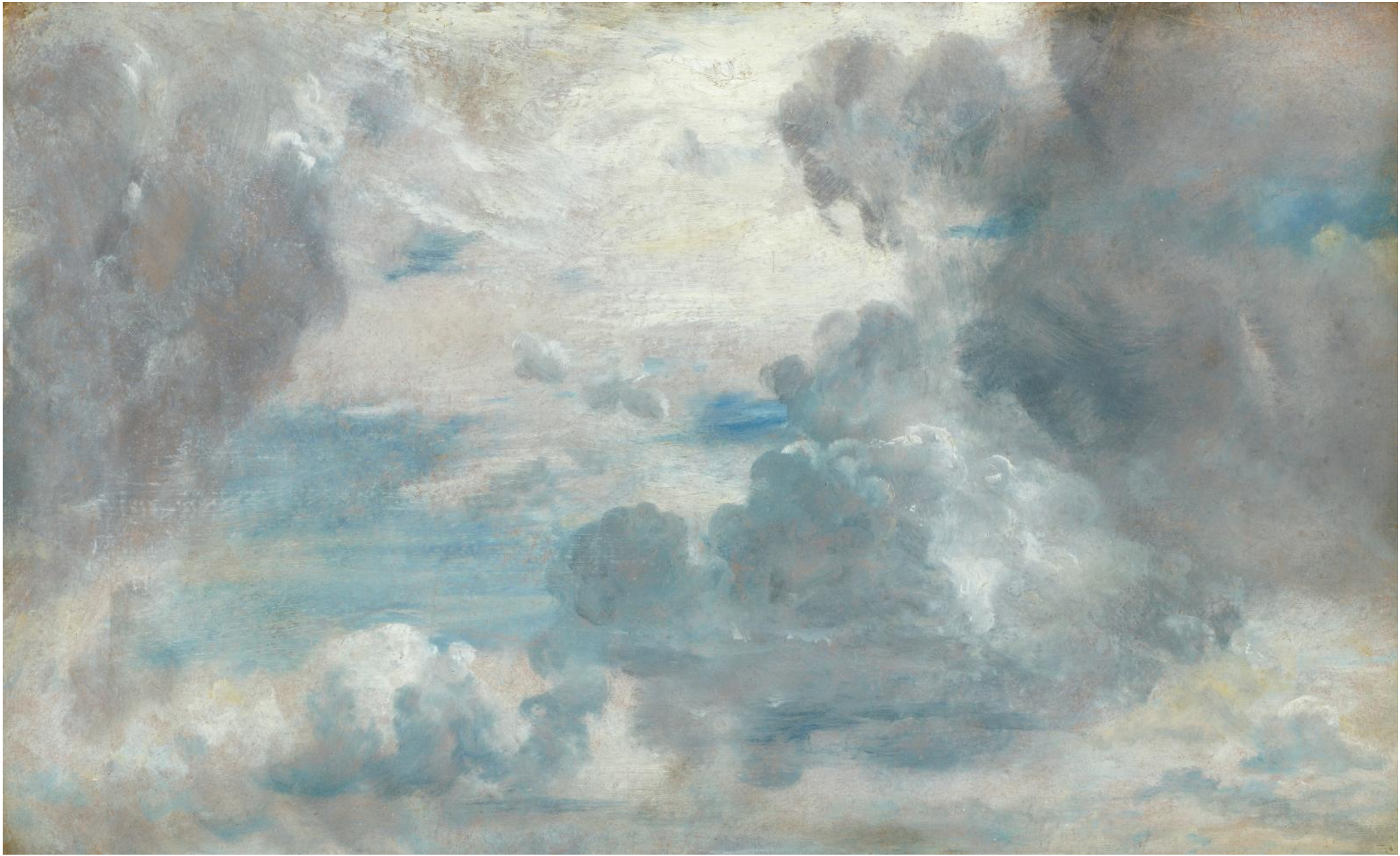}
  \label{fig:sub2}
\end{subfigure}
\vspace{-0.1in}\caption{Two {\it Cloud Study} oil paintings by John Constable (1822). Left: Yale Center for British Art.
Right: The Frick Collection.
}\label{fig:rep}
\end{figure}

\subsection{The Art-Historical Questions}
In 1821, Constable undertook a sustained campaign of ``skying,'' as he called his outdoor sketching of clouds. There is general art-historical agreement that Constable's painted clouds became more life-like around this time (Fig.~\ref{fig:rep})~\cite{thornes1999john, lyles2004glorious}. The significance of this period of concentrated effort has been debated~\cite{lyles2004glorious,cove2004great,gage2000cons}. Some see Constable's cloud paintings of this period as confirmation that the artist's powers of observation improved as a consequence of prolonged study, enabling him to execute more convincing clouds~\cite{reynolds2004constable}. Yet faithful visual documentation of clouds is challenging because they are {\it constantly changing}. It seems reasonable to posit that Constable relied on certain artistic conventions or formal patterns for his paintings of these ever-shifting motifs, as painters often did. It has also been argued that the 1821 skying campaign was a belated response to the 1803 publication of Luke Howard's typology of clouds into cumulus, cirrus, stratus, etc.~\cite{howard1803modi}, though there is no direct evidence that Constable consulted Howard's publication~\cite{badt1950john}. Scholars remain in disagreement about the degree to which Constable relied strictly on empirical observation, on visual formulae that might escape the notice of human viewers, or on a new understanding of how to distinguish and thus represent different types of clouds~\cite{lyles2004glorious,cove2004great,gage2000cons}. To some extent, scholarly disagreement arises from the fact that human viewers may not perceive or may perceive only with difficulty qualities like 
cloud accuracy or visual conventions that have been naturalized through regular use by European artists. Our goal is to contribute to a more accurate understanding of Constable's realism via three paths of inquiry: 
\begin{enumerate}
    \item 
    Do Constable's clouds correspond with the system of cloud typology introduced in 1803 by Luke Howard? 
    \item
    How closely do Constable's paintings 
    emulate the appearance of actual clouds when compared to photographs of clouds?
    \item
    How does the empirical accuracy of Constable's clouds compare with that of his contemporaries when judged against photographs of clouds?
\end{enumerate}

\subsection{Overview of Our Approach}\label{sec:overview_approach}

\begin{figure}[ht!]
  \centering
  \includegraphics[width=\linewidth]{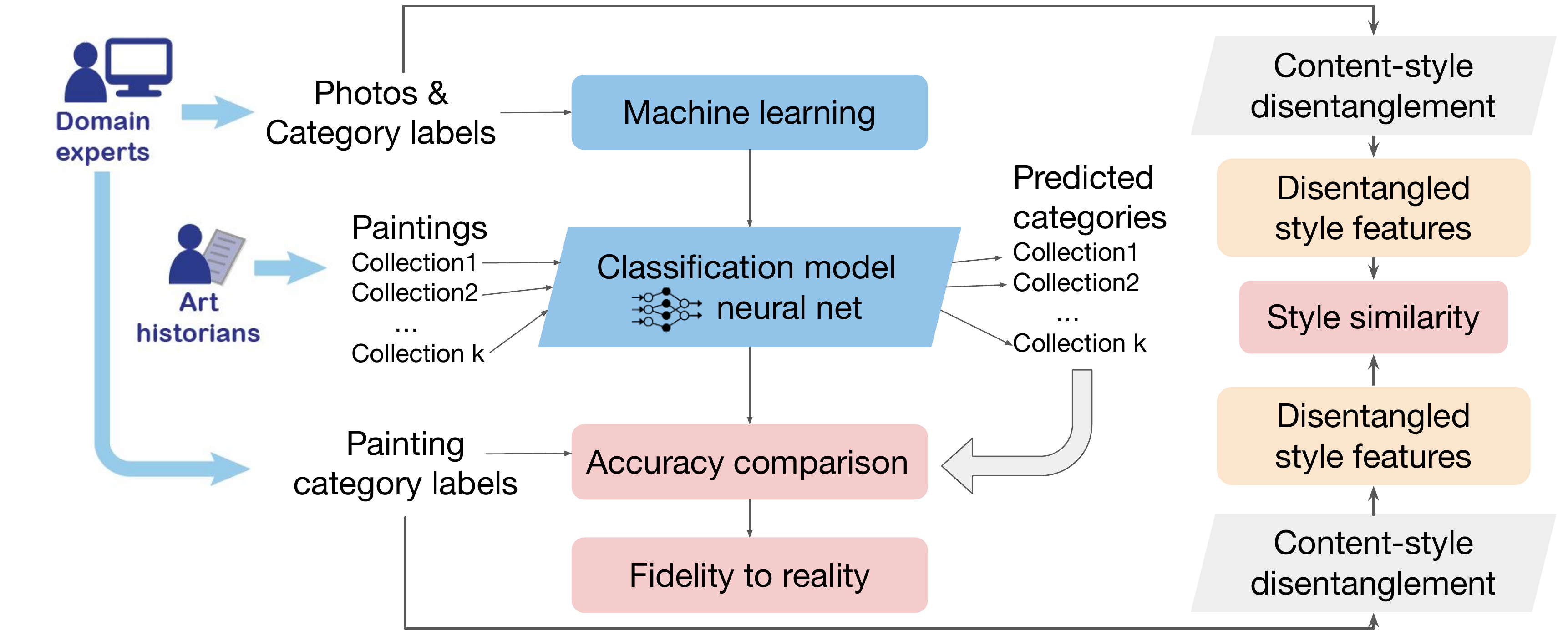}\vspace{-0.1in}
  \caption{The proposed machine learning paradigm for studying pictorial realism.}
  \label{fig:ml}
\end{figure}

These judgments about realism from art historians are highly subjective insofar as they record the opinion of a particular viewer at a particular moment. The perceived fidelity of a painting to the natural phenomena it represents cannot always be clearly explained, because it is guided by an immediate, intuitive response to a particular painting. This is especially true of hard-to-describe phenomena like clouds or crashing waves: for most human viewers, paintings of these subjects simply ``look right'' or not. To provide a more objective assessment of realism, we introduce a machine-learning-based analysis procedure. As shown in Fig.~\ref{fig:ml}, this method comprises two components: classification of painted content (cloud in this case) and evaluation of painting style. In a nutshell, we evaluate pictorial realism by assessing the similarity between paintings and photographs in terms of both the painted 
content and painting style, which makes our evaluation system more thorough and unbiased~\cite{zhang2022reducing}.

After obtaining a labeled dataset containing both photographic images of clouds and a collection of sky paintings, we first train a machine learning system using these photographs to  classify cloud categories. We then apply this classifier to our painting set to predict their cloud categories. In the meantime, classification labels are created for the paintings by experts (meteorologists). The classification accuracy for the paintings is then computed and compared with the accuracy achieved for photographs. Our basic assumption is that the classification accuracy of paintings that imitate observed reality well will be close to that obtained for the photos. Further comparison can be conducted between different collections of paintings, allowing assessment of the relative fidelity of various collections to nature. One type of comparison across collections is between works by different artists. Our labeling relies on the expertise of meteorologists to categorize clouds documented in photographs and paintings according to the types defined by Howard~\cite{howard1803essay}. We propose a semi-supervised learning model for cloud classification that merges classic features with edge features. The classification of clouds in Constable's paintings according to the standard typology allows for a more precise comparison with his contemporaries. By contrasting the AI system's predictions with the expert-created ground truth labels, we obtain an objective assessment of the degree to which painters are (knowingly or unknowingly) differentiating cloud types. Given the highly specialized skills and knowledge required to classify cloud types, the AI system offers an insight unattainable by the average human viewer. 

Furthermore, to further explore painting styles, we examine pictorial realism from another perspective of painting style. Specifically, we first extract the encoded style features from each painter's collection by training a content-style-disentanglement model~\cite{huang2018multimodal}. Using our newly developed evaluation metrics, we assess the pictorial realism based on these extracted style features. This allows us to compare the relative realism of various painting styles in our dataset against that of John Constable. These style features act as direct representations of the unique pictorial characteristics of each painter's collection in comparison to photographic images.

The {\bf key contributions} of our work include:
\begin{itemize}
    \item {\it Interdisciplinary framework:}
    We proposed a machine learning framework to study realism in art from an explainable and interdisciplinary perspective by leveraging computer vision techniques, meteorology expertise, and art history insights. 
    \item {\it Methodology:}
    We developed several tools and models: a sky-ground segmentation algorithm, a new semi-supervised CNN model (named SFF-CNN) for cloud-type classification, and new evaluation metrics to quantify the style differences between images. Notably, this is the first effort to harness unlabeled sky photos to enhance cloud classification.
    \item {\it Dataset:}
    We curated a unique dataset consisting of 363 paintings featuring skies by John Constable and six of his contemporaries. Two expert meteorologists professionally annotated each piece, making it the inaugural dataset of paintings designed for computational analysis of skies. We are sharing our sky segmentation results and detailed annotations with the broader research community.  
    \item {\it Insights:}
    Our findings furnish the art history domain with compelling evidence:
    {\it Constable's systematic adherence to cloud typologies is pivotal for the pronounced realism in his cloud artworks.}
\end{itemize}


\subsection{Related Work} \label{sec:related}

We briefly introduce related work on the art-historical study of Constable's sky paintings, computerized cloud-type classification, and content-style disentanglement.
Modern art-historical scholarship on Constable's clouds began with Kurt Badt's 1950 book on the subject~\cite{badt1950john}. Prior to this, accounts of Constable's clouds were largely descriptive as opposed to analytical, attributing their realism to Constable's emotional connection with nature, his devotion to sketching outdoors, or his largely rural childhood~\cite{chamberlain1903john}. Badt was the first to argue that Constable's proficiency with painting realistic clouds was due to his familiarity with the recent development of a typology of clouds created by British chemist Luke Howard.  Howard's typology was published in 1803 and was widely disseminated during Constable's lifetime, so it was available to him. But there is no evidence that Constable possessed Howard's typology, and the artist's extant correspondence makes no direct reference to Howard~\cite{hawes1969constable}. More recent scholars tend to cite instead Constable's dedication to sustained periods of empirical observation of clouds~\cite{thornes1999john,reynolds2004constable} and his familiarity with earlier paintings of naturalistic landscapes by artists like Claude Lorrain or Willem van de Velde the Younger, both of whom were well represented in English art collections during Constable's lifetime~\cite{hawes1969constable,esmeijer1977cloudscapes}. In addition, a Romantic explanation for Constable's naturalism likewise persists in the scholarly literature to this day, attributing his naturalism at least in part to an emotional or spiritual impulse toward accuracy in his depictions of natural phenomena~\cite{lyles2004glorious}.
 
We regard the accuracy of cloud-type classification as strong evidence of Constable's familiarity with Howard's typology, so building a trustworthy cloud-type classifier is indispensable. 
 Recently, researchers have started to adopt CNNs for cloud-type classification. Zhang {\it et al.}~\cite{zhang2018cloudnet,zhang2020ensemble} built a large ground-based cloud dataset, called Cirrus Cumulus Stratus Nimbus (CCSN) with cloud type labels, and a CNN model for cloud classification. Huertas {\it et al.}~\cite{huertas2020cloud} proposed a feature fusion model combining CNN features and handcrafted low-level textural features to boost classification accuracy. Departing from this fusion model, our approach aims to extract more task-relevant features such as the contours of clouds to improve classification on both the photo and painting datasets.
 
 Another problem that we address is the lack of labeled cloud photos. The emergence of semi-supervised learning can enhance classification performance by utilizing a great amount of unlabeled data during the training process. The common semi-supervised classification models can be categorized into self-learning~\cite{sohn2020fixmatch}, co-training~\cite{fan2022ucc}, graph-based semi-supervised learning~\cite{song2022graph}, and semi-supervised supported vector machine~\cite{wan2021semi}. Following the idea of self-learning, we generate pseudo labels (detailed in Section~\ref{sec:ssl}) for two unlabeled sky photo datasets and then add these new data to the labeled CCSN dataset to achieve dataset expansion.
 
 Content-style disentanglement has been extensively applied for feature decoupling, with both the content and style feature representations useful for downstream problems, such as semantic segmentation~\cite{chartsias2019disentangled,liu2021semi}, image retrieval~\cite{sain2021stylemeup,lee2021cosmo}, and image style transfer~\cite{kwon2021diagonal,kotovenko2019content}. In image translation, most CNN-based methods aim to learn latent space representations by extracting content or style information using autoencoder variants. However, utilizing these disentangled features for similarity or discrepancy comparison among paintings from different artists--as we have done in this study--is a relatively uncharted territory. 
 


\begin{figure}[ht!]
\centering
\begin{tabular}{m{0.45\linewidth}m{0.45\linewidth}}
\centering
        \includegraphics[width=0.98\linewidth]{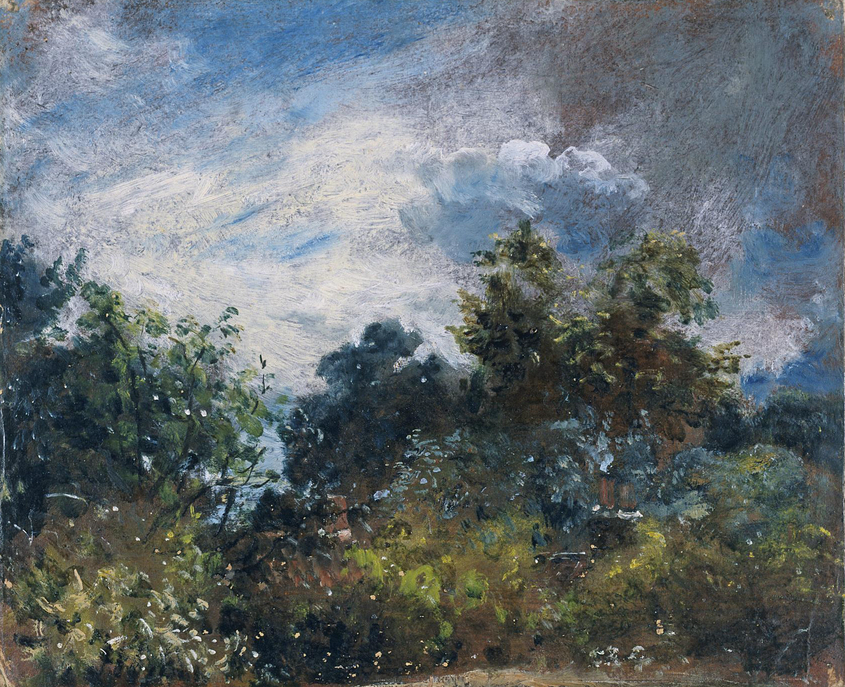} &
 {%
\setlength{\fboxsep}{0pt}%
\setlength{\fboxrule}{1pt}%
       \fbox{\includegraphics[width=0.98\linewidth]{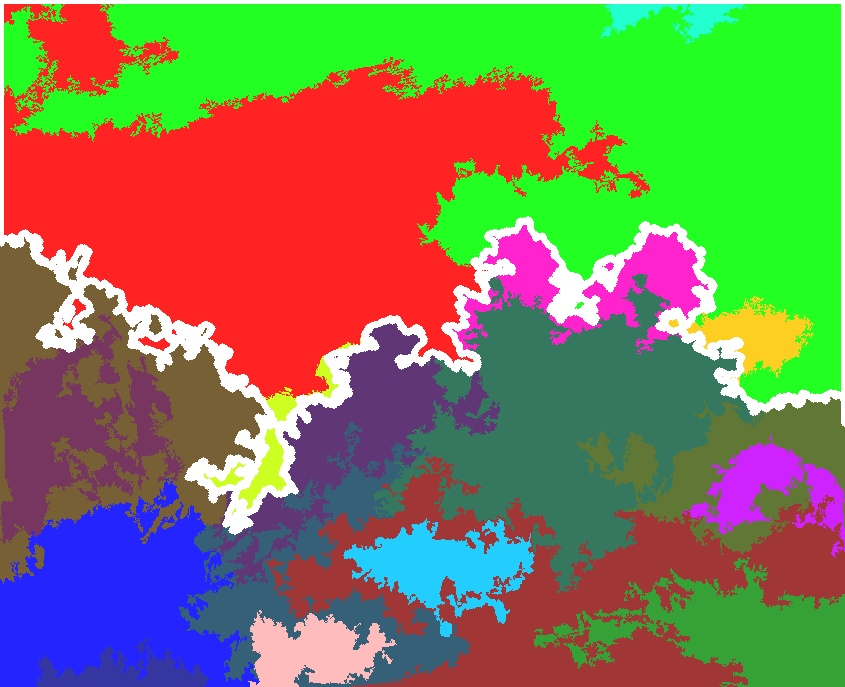}}
       }%
\end{tabular}\\
        (a) John Constable, {\it Study of Sky and Trees}, 1821\\ \vskip 0.05in
        
\begin{tabular}{m{0.45\linewidth}m{0.45\linewidth}}
\centering
        \includegraphics[width=0.98\linewidth]{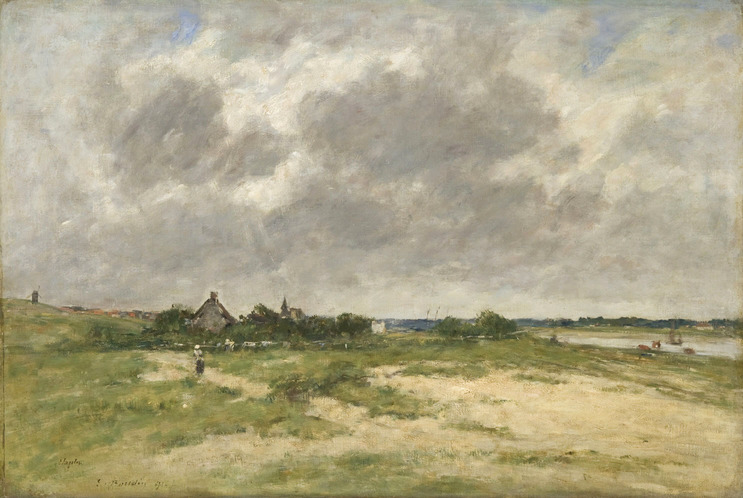} &
 {%
\setlength{\fboxsep}{0pt}%
\setlength{\fboxrule}{1pt}%
        \fbox{\includegraphics[width=0.98\linewidth]{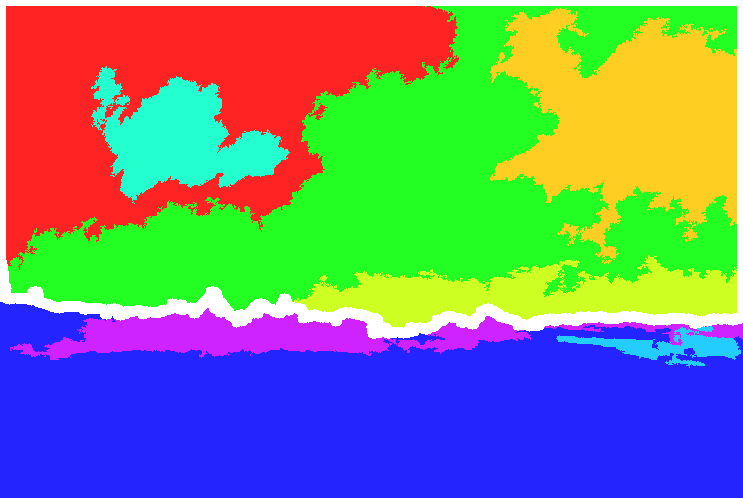}}
        }%
        \\
\end{tabular}\\
 (b) Eug\`ene Boudin, {\it Etaples, les Bords de la Canche}, 1891 \vspace{-0.1in}

\caption{Sky and ground segmentation illustrated with two paintings. Left: Original paintings. Right: Homogeneous patches (represented by different colors) generated using the A3C algorithm.} Regions within the thin white contours are the sky regions after regression.
\label{fig:seg}
\end{figure}

\section{Algorithms}\label{sec:ml}

As we have discussed in Section~\ref{sec:overview_approach}, our paradigm for studying pictorial realism (Fig.~\ref{fig:ml})
provides a novel perspective for comparing artworks with photographs and addresses the subjectivity of experts' opinions. Below, we elaborate on the technical components in the analysis pipeline.

\subsection{Semi-Supervised Cloud-Type Classification}

Our classification model consists of two main steps: clustering-based sky segmentation and classification by a {\it semi-supervised feature fusion CNN} (SFF-CNN) model. The sky segmentation step reduces the impact of irrelevant parts of an image on classification. SFF-CNN contains two streams of feature extraction, aptly called the {\it classic feature extractor} and {\it edge feature extractor}. The former generates features from low-level textures or patterns to high-level object-related characteristics, while the latter focuses on edge information. The fused features from the two encoders are utilized together for the ultimate class label prediction. We are motivated to extend a typical CNN model by incorporating edge features because (1) the contour information of cloud bases and updraft turrets is valuable for meteorologists to determine the cloud type, and (2) 
CNN models tend to focus on texture rather than shape for recognition~\cite{geirhos2018imagenet} while paintings and photos have different texture characteristics.
Our extended CNN model is trained iteratively by generating pseudo labels for unlabeled images and then refitting the model.


\subsubsection{Sky Segmentation}
The land, mountains, or other irrelevant regions in a painting can negatively affect cloud classification. Because only sky regions are used in the training photos, we eliminate the impact of other irrelevant parts in the paintings by excluding pixels outside the sky region from subsequent classification analysis. Specifically, a painting is segmented into two classes: sky versus non-sky (mostly land). The entire non-sky region of a painting image is replaced by black pixels and the modified image becomes the input to the CNN model, which we refer to as the {\it sky-selected image}.

Our sky segmentation algorithm includes two major steps: segmentation into homogeneous patches (aka, segment) and classification of each segment into sky versus non-sky. For the first step, we used the Agglomerative Connectivity Constrained Clustering (A3C) algorithm~\cite{li2011agglomerative}. 
For the second step, we perform logistic regression on the features extracted from each segment to determine whether the segment is sky or non-sky. For each segment, a 10-dimensional feature vector including location and color-based features is computed. Details about the sky detection algorithm and some example results are provided in Supplementary Materials.
Fig.~\ref{fig:seg} shows the clustering results of two paintings and the sky versus non-sky classification results of the segments.

\subsubsection{Cloud-Type Classification}
The sky-selected images are classified into different cloud types by the SFF-CNN model. Our neural network is custom-designed for cloud-type classification by incorporating pre-learned edge features into the layers of a typical CNN model as edge information is crucial in differentiating various types of clouds. The neural network consists of a bottom stream for classic feature extraction and a top stream for edge feature extraction. The classic feature extractor aims at extracting useful features from low-level textures or patterns to high-level object-related quantities, while the edge feature extractor only captures the characteristics of edges in the same input image. Both feature extractors take the three-color-channel sky-selected images as the input. 

{\bf Classic Feature Extraction:}
Denote the $k$th sky-selected image by $\mathbf{I}_k^{}$. The encoder for classic feature extraction takes the three-color-channel image $\mathbf{I}_k$ as the input.
The first two convolutional blocks both consist of two Conv-BatchNorm-Relu layers and are followed by a $2\times2$ pooling layer to downsample the input feature maps ($400 \times 400$). The convolutional layers in these two blocks all have stride set to $1$ and the kernel size $3\times3$. The next two blocks are residual blocks with two convolutional layers with stride set to $1$ and $2$, respectively, and the same kernel size $3\times3$. Each of these blocks spatially downsamples the input feature maps to half of their size. 
The third residual convolutional module follows the same structure as the first two but sets stride to $1$ for both convolutional layers. Then three fully-connected layers with feature dimensions 4096, 1024, and 10, respectively, are connected to the Resconv modules. The final layer of Softmax activation produces a distribution over the ten output probability classes for each category. Lastly, the cross-entropy (CE) loss~\cite{murphy2012machine} is applied to train the network.


{\bf Edge Feature Extraction:}
Visualization results using the Grad-cam method~\cite{selvaraju2017grad} (shown in Supplementary Materials) verified our expectation that edge information is important for classifying cloud types, which motivated our strategy to fuse edge features in the CNN. We compute the edge features by a pre-trained encoder named holistically-nested features for edge detection (HED)~\cite{xie2015holistically}.  The side-output layer of each convolution module of HED generates an edge feature map at a particular receptive field size.
These maps are concatenated with those generated by the CNN at corresponding layers. The two feature maps are ensured to have the same size (horizontally and vertically) such that the features at any location on one map can be combined with features at the same location on another map before convolution.
In particular, we use the same setting for the HED and CNN architectures so that at every layer, their respective feature maps are generated with the same receptive field size. The augmented feature map is the input to the next convolution layer.

\subsubsection{Semi-Supervised Learning}\label{sec:ssl}
To further enhance cloud classification accuracy, we employ semi-supervised learning to leverage a large set of 9,883 unlabeled cloud photos from the SkyFinder dataset~\cite{mihail2016sky} and FindMeASky dataset~\cite{rawat2018find}. We also apply data augmentation following the schemes of FixMatch~\cite{sohn2020fixmatch}. For each unlabeled image, its flipped and shifted versions, called weak augmentation images, are created.  Additionally, the so-called strong augmentation images are created by another two operations, namely, CTAugment followed by Cutout~\cite{sohn2020fixmatch}. We first apply the classifier trained using only the labeled images to classify the weak augmentation images. The class that has the maximum predicted posterior probability is chosen as the predicted class (also called the one-hot pseudo label). To counter the negative effect of possibly incorrect pseudo labels, the maximum predicted posterior is compared with a pre-chosen threshold. If the threshold is not exceeded, this unlabeled image and its augmented versions will not be used further.  Otherwise, the pseudo label is treated as the true label for the strong augmentation images, which we refer to as high-confidence unlabeled images. Finally, another round of training is performed using both labeled and high-confidence unlabeled images. The cross-entropy between the true class and the labeled images and between the pseudo-class generated from the weak-augmented images and the predicted class posteriors using the strong-augmented images are defined as the loss to train the model. 
\subsection{Style Disentanglement}
In addition to comparing paintings based on how well they can be classified, we propose a methodology to 
assess the similarity in the ``style'' features of pictures. In MUNIT~\cite{huang2018multimodal}, an image is decomposed into two representations: content versus style. Both the content and style features are extracted by an encoding CNN, and they can be combined as input to a decoding CNN to reconstruct the original image. Roughly speaking, the content features capture the shared characteristics between two sets of images, whereas the style features pinpoint the unique attributes of each set. The encoders and decoders for both image sets are trained together to ensure that the content features correspond to traits shared by the two sets. 


In our analysis, we treat the set of paintings of every artist as domain $\mathcal{A}$ and the set of cloud photographs as the reference domain $\mathcal{P}$. This training process yields a content encoder and a style encoder for each artist. The training algorithm generates photo-realistic images $I_{\mathcal{A}2\mathcal{P}}$ from images in domain $\mathcal{A}$ or painting-like images $I_{\mathcal{P}2\mathcal{A}}$ from those in domain $\mathcal{P}$, an operation called ``cross-domain style translation.'' The translation is achieved by keeping the content features but adopting style features generated for an image in the other domain. These cross-domain features are fed into a decoder to reconstruct a translated picture. The training objective function used in~\cite{huang2018multimodal} has been modified slightly in~\cite{liu2020metrics} by removing the learning regression loss because the authors of the latter found that better separation of content and style can be obtained and the style and input image will be more correlated. In subsequent discussions, we will refer to the style features computed via a style encoder simply as the ``style'' of an image.


\subsubsection{Style Similarity Between Artists} First, to evaluate style similarity between artists, we consider two sets of paintings denoted by $A$ and $B$. Suppose $A=\{ a_i : i \in \{1,2,...,n_A\}\}$ contains $n_A$ pictures and $B=\{b_j : j \in \{1,2,...,n_B\}\}$ contains $n_B$ pictures. Denote the content and style encoder trained based on style transfer from painting set $A$ to photo set $P=\{ p_k : k \in \{1,2,...,n_P\}\}$ by $E_C^{A}$ and $E_S^{A}$, respectively. Likewise, the encoders for ${B}$ are $E_C^{B}$ and $E_S^{B}$. For an image $a_i\in {A}$, denote its style features computed by $E_S^{A}$ by $F_S^{a_i}$. Similarly, for any $b_j\in {B}$, let its style computed by $E_S^{B}$ be $F_S^{b_j}$.
If ${A}$ and ${B}$ are similar in style, we would expect $F_S^{a_i}$ and $F_S^{b_j}$ to be close on average. Use the normalized square of the $L^2$ norm of a style feature vector to indicate the signal strength: $I_S^{a_i}=\|F_S^{a_i}\|^2/d$, where $d$ is the dimension of the style feature vector. The Mean Squared Error (MSE) between $F_S^{a_i}$ and $F_S^{b_j}$ is simply $\|F_S^{a_i}-F_S^{b_j}\|^2/d$.
For each image $a_i \in {A}$, we define its average distance to images in ${B}$ by 
\begin{eqnarray}
D_{A}^{a_i}=\frac{1}{n_B}\sum_{b_j\in {B}}\frac{\text{MSE}(F_S^{a_i}, F_S^{b_j})}{I_S^{a_i}}\;.
\end{eqnarray}
Conversely, for each image $b_j\in {B}$, we define its average distance to images in ${A}$ 
as $D_{B}^{b_j}$ likewise.
Finally, define $D_{{A}}=\frac{1}{n_A}\sum_{a_i\in {A}}D_{A}^{a_i}$, 
$D_{{B}}=\frac{1}{n_B}\sum_{b_j\in {B}}D_{B}^{b_j}$, and 
\begin{eqnarray}
D_{\text{style}}({A},{B})=\frac{1}{2}(D_{{A}}+D_{{B}}).
\label{e3}
\end{eqnarray}
The distance $D_{\text{style}}$ is taken to measure the style difference between sets ${A}$ and ${B}$.


\subsubsection{Style Similarity Between an Artist and Photos} 
Next, we 
propose to use the metric ``Information Over Bias (IOB)''~\cite{liu2020metrics} to measure the difference between the paintings of an artist and real photos. For an image $a_i \in {A}$, where $a_i$ is treated as a vector, let its style feature vector be $F_S^{a_i}$.
$\text{IOB}(a_i, F_S^{a_i})$ is defined to quantify the amount of information in $a_i$ which is captured by $F_S^{a_i}$. Specifically, the informativeness of $F_S^{a_i}$ is measured by the ratio between $\text{MSE}(a_i,\widetilde{a_i}')$ and $\text{MSE}(a_i,\widetilde{a_i})$, where $\widetilde{a_i}'$ is a reconstructed image from an uninformative constant substitute style vector $\mathbbm{1}$ combined with $a_i$'s content feature vector, while $\widetilde{a_i}$ is generated from the informative style vector $F_S^{a_i}$ and the same content vector. 
Thus, we define $\text{IOB}(a_i, F_S^{a_i})$ by
 $\text{IOB}(a_i, F_S^{a_i})=\text{MSE}(a_i,\widetilde{a_i}')/\text{MSE}(a_i,\widetilde{a_i})$.
With a slight abuse of notation, we also use $\text{IOB}({A})$ to denote the average $\text{IOB}$ values for the images in ${A}$, i.e.,
 $\text{IOB}({A})=\frac{1}{n_A}\sum_{i=1}^{n_A} \text{IOB}(a_i, F_S^{a_i})$.
A lower value of $\text{IOB}({A})$ indicates that the style representation of the image is less important since a substitute default style vector can result in reconstruction with a similar level of disparity from the original image. Because the style feature vectors capture the distinct characteristics of one set of images from another set, less informative style vectors reflect a higher similarity between the two set of images. To form a basis of comparison, we also compute $\text{IOB}$ for a mixed set containing both paintings and cloud photographs. Specifically, we first compute $\text{IOB}({A})$ for a set of paintings by an artist using the style transfer process from paintings to photographs. Then we mix images from the painting set ${A}$ and the photo set ${P}$ to form a new set ${M}=\{a_i: i\in \{1,2,..., n_A\},~p_k: k \in \{1,2,..., n_P\}\}$. Again by the style transfer process from set ${M}$ to ${P}$, we can compute $\text{IOB}({M})$. Finally, the style distance between an artist and the photographs is defined as $\displaystyle R_\text{style}({A})=\text{IOB}({M})/\text{IOB}({A})$.

\section{Experimental Results}\label{sec:exp}

\subsection{Painting and Photo Datasets}
We curated a dataset of oil paintings by John Constable (1776-1837) and six of his near-contemporaries: Pierre Henri de Valenciennes (1750-1819), David Cox (1783-1859), Frederick Richard Lee (1798-1879), Frederick W. Watts (1800-1870), Eugène Boudin (1824-1898), and Lionel Constable (1828-1887). All of these images are either high-resolution scans of existing reproductions or digital photographs of landscape paintings with ``finished'' clouds or pure cloud studies. 

Cloud types and detailed meteorological information for each painting in the dataset were labeled by two meteorologists with expertise in cloud classification. One annotator possesses basic knowledge of the history of European landscape painting, while the other does not. Post their initial round of labeling, the two experts reached consensus on 75.5\% of the labels. They both recognized that the majority of different annotations were due to borderline cases. Following a discussion between the experts, the labels used in the subsequent experiments were mostly based on the senior annotator's annotations, while the labels of 15 paintings were in accordance with the junior annotator's opinion. 
Finally, an open dataset containing 363 images with detailed labeled metadata was established, which will be shared (to the extent that image licensing allows) in order to facilitate further analyses of the relation between painted clouds and actual meteorological phenomena.


We used the CCSN dataset to train the cloud classification model. The CCSN dataset contains 2,543 cloud images, in which cloud photographs were labeled into 10 cloud categories, thus we formulated cloud-type classification as a 10-class problem. For semi-supervised learning, we leveraged the SkyFinder~\cite{mihail2016sky} and FindMeASky~\cite{rawat2018find} datasets, which came with the sky segmentation masks but no cloud-type labels. After eliminating duplicate images, our unlabeled dataset comprised 9,883 photos.

\subsection{Cloud Classification on the Paintings}
To evaluate our sky segmentation algorithm, we manually labeled sky regions for all 363 paintings, which serve as the ground truth. We then computed pixel accuracy, mean accuracy, and mean IoU as evaluation metrics, which were 0.9804, 0.9613, and 0.9427, respectively. Such accuracy levels are regarded as high.

\begin{figure}[ht!]
  \centering
  \includegraphics[width=0.98\linewidth,trim={40 20 58 36},clip]{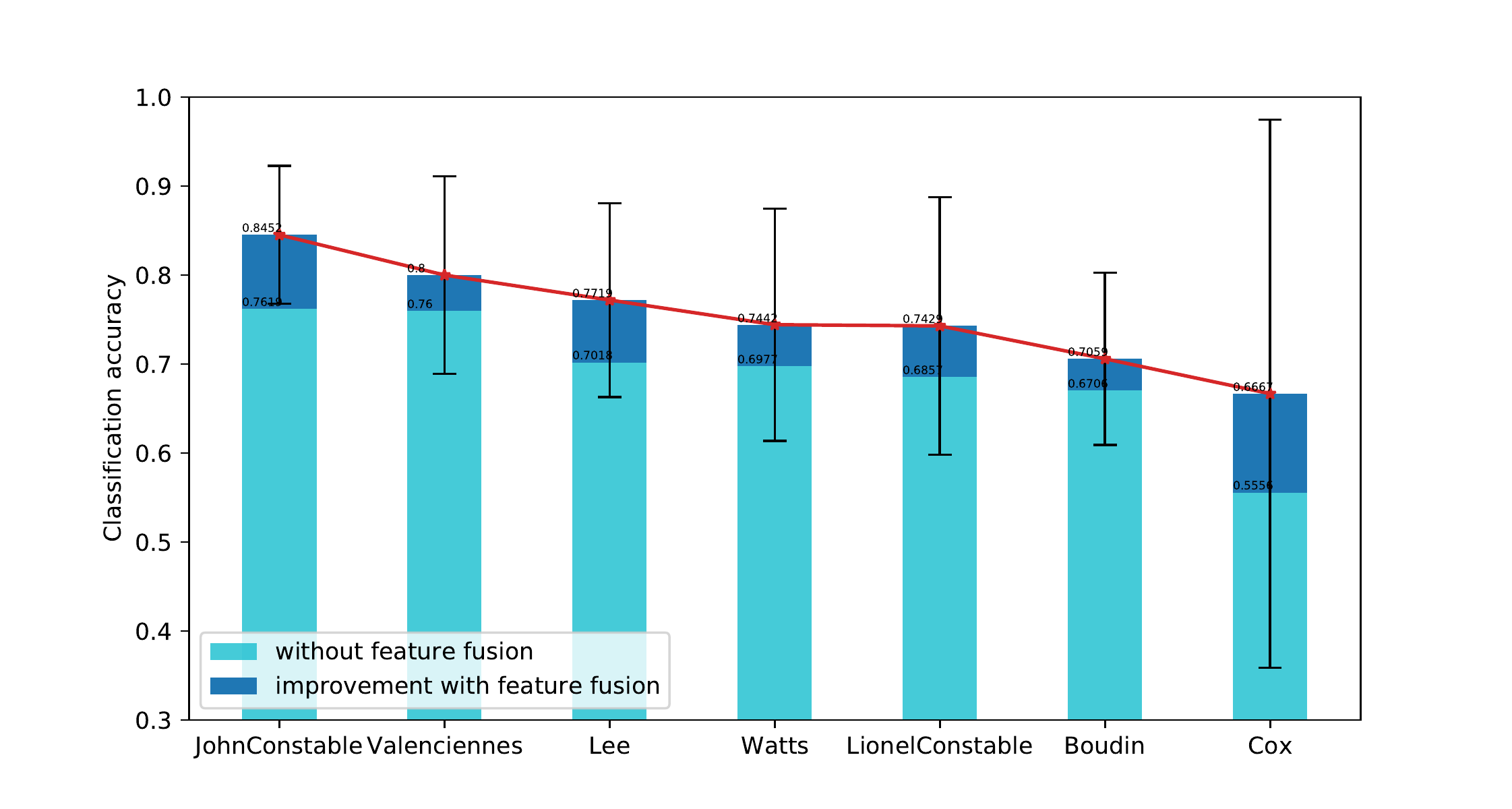}
  \vspace{-0.1in}
  \caption{The comparison in terms of classification accuracy of all seven painters using the SFF-CNN model with or without feature fusion. The error bars denote the confidence interval for the classification accuracy at the significance level of 0.05 for each painter.}
  \label{fig:bar_result}
\end{figure}

\begin{figure*}[ht!]
\setlength{\tabcolsep}{0.8pt}
\centering
\begin{tabular}{ccccccc}
        \includegraphics[height=1in]{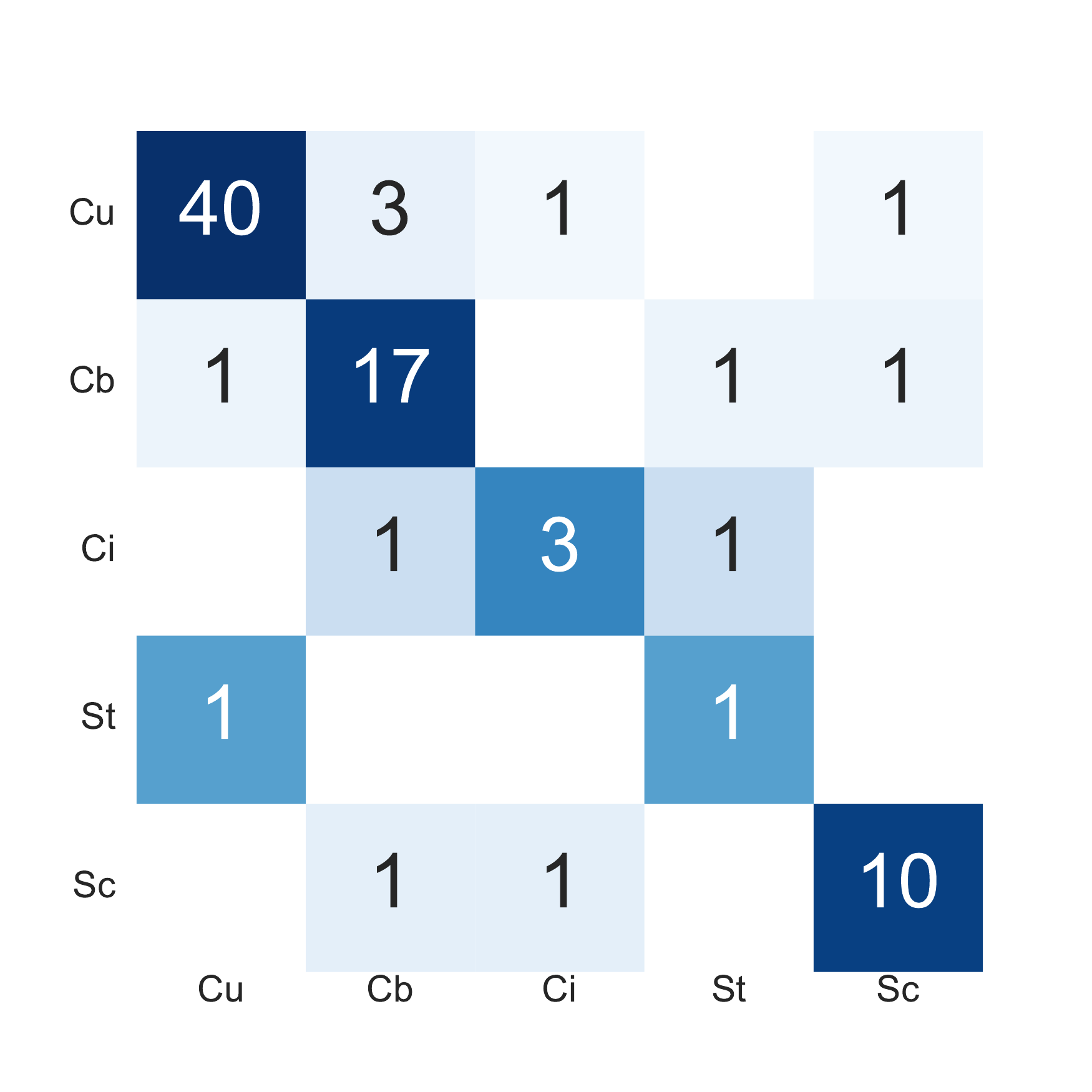} &
        \includegraphics[height=1in]{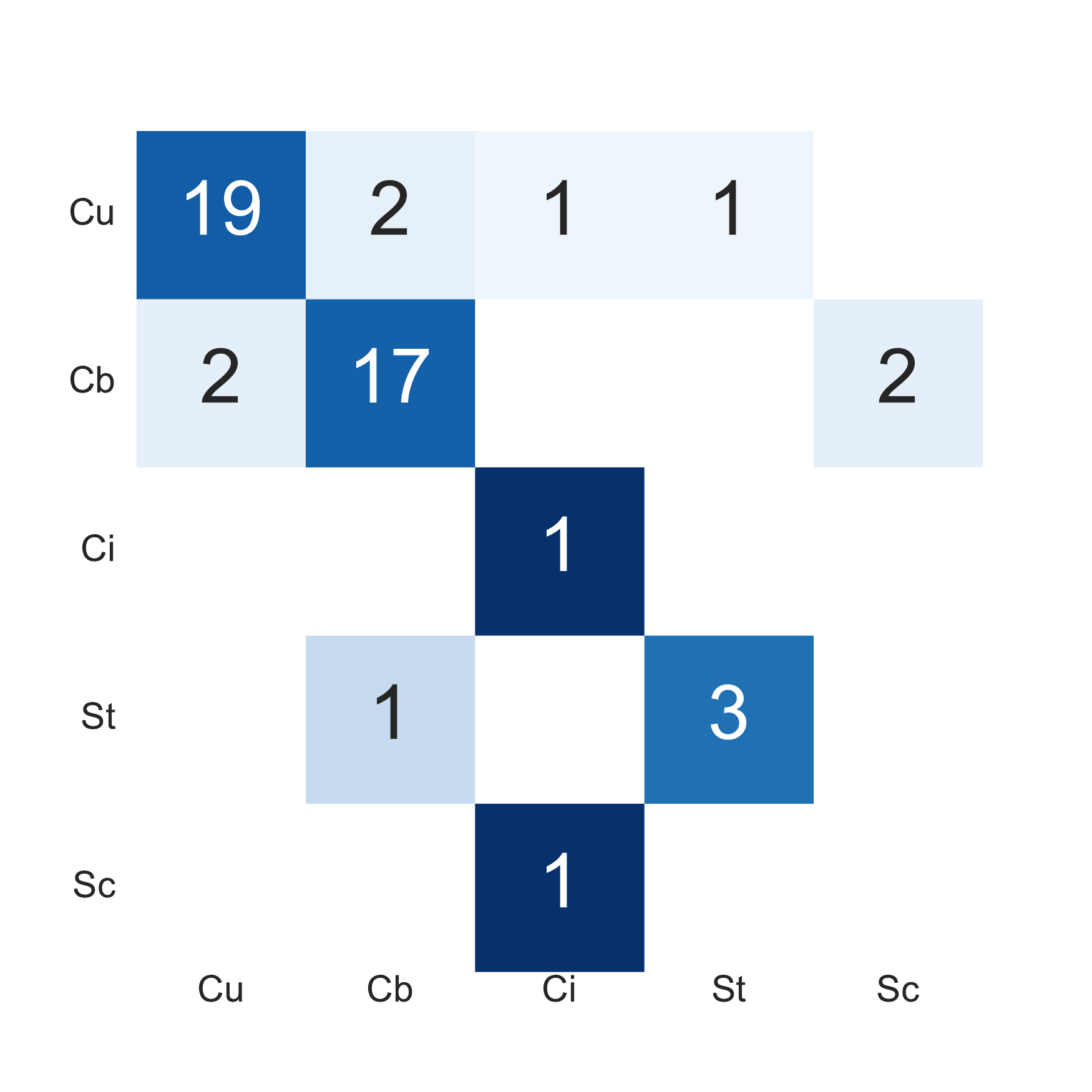} &
         \includegraphics[height=1in]{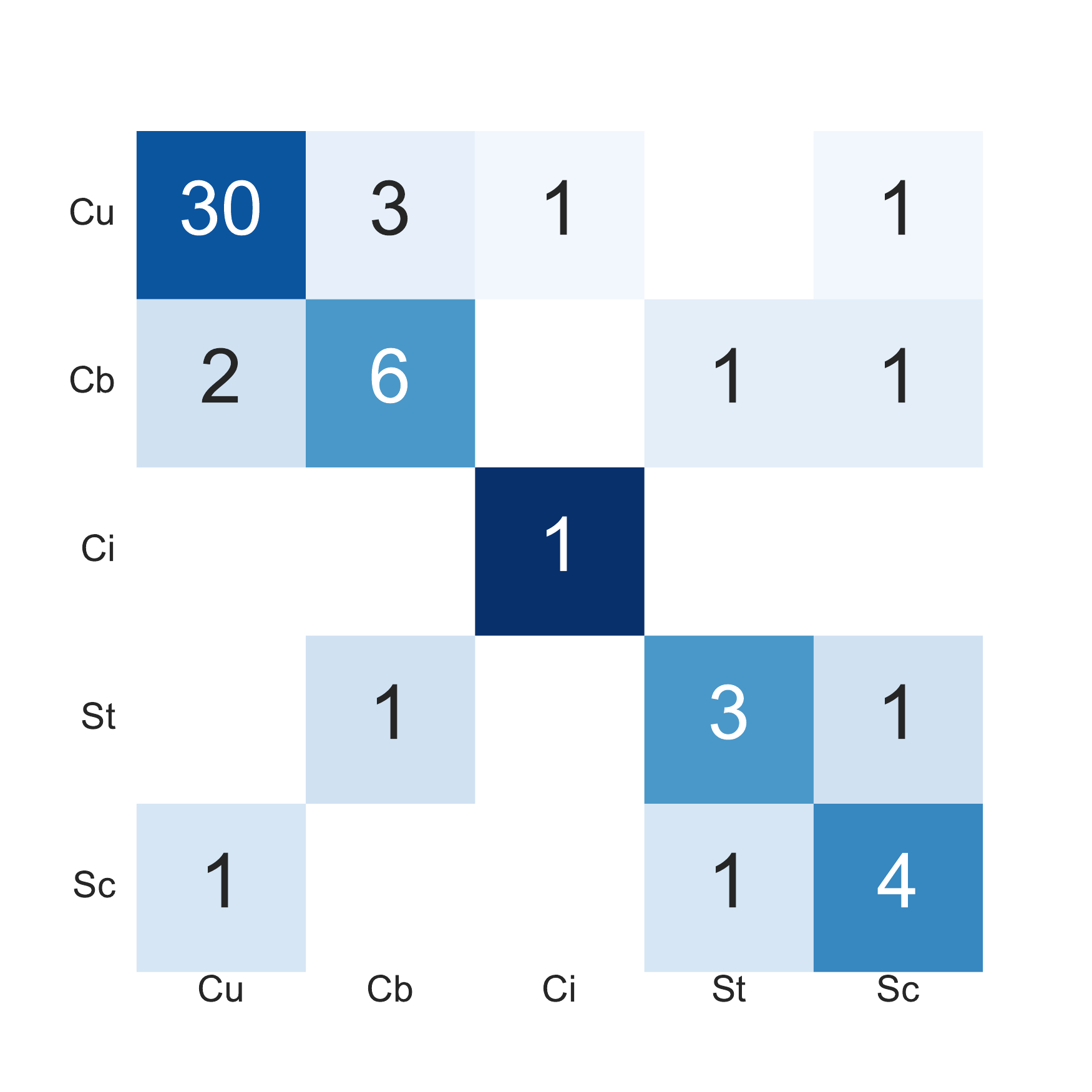} &
        \includegraphics[height=1in]{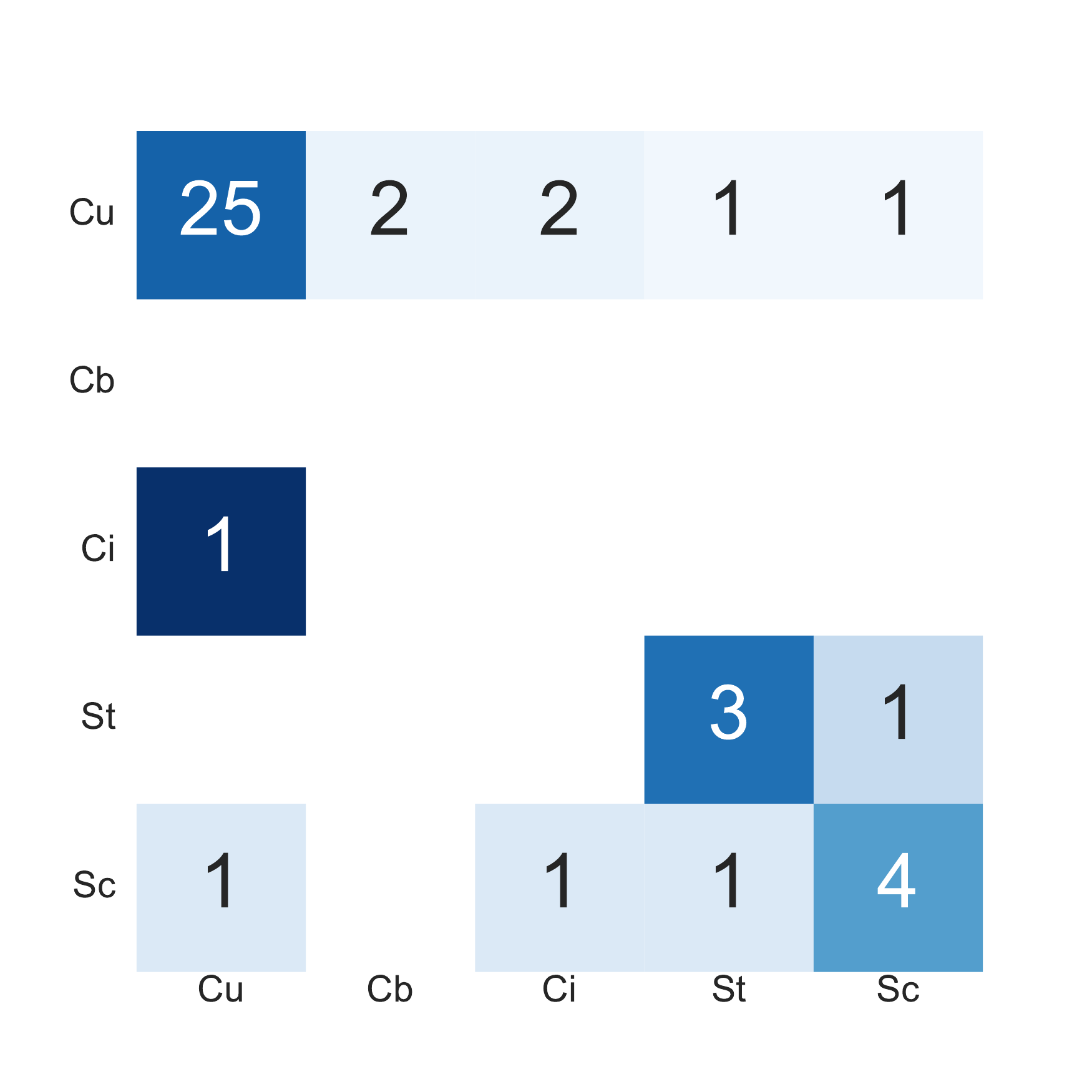} &
       
         \includegraphics[height=1in]{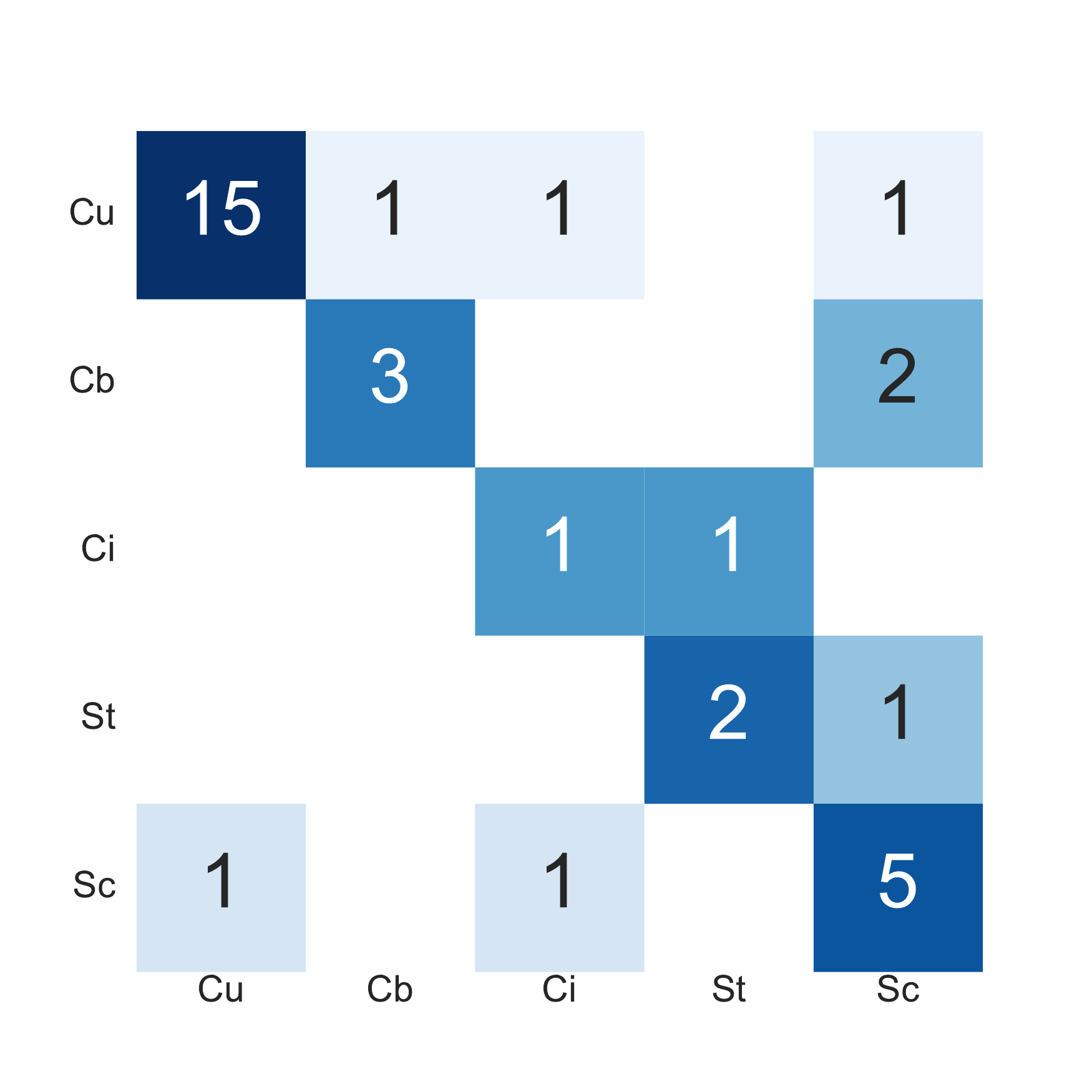} &
        \includegraphics[height=1in]{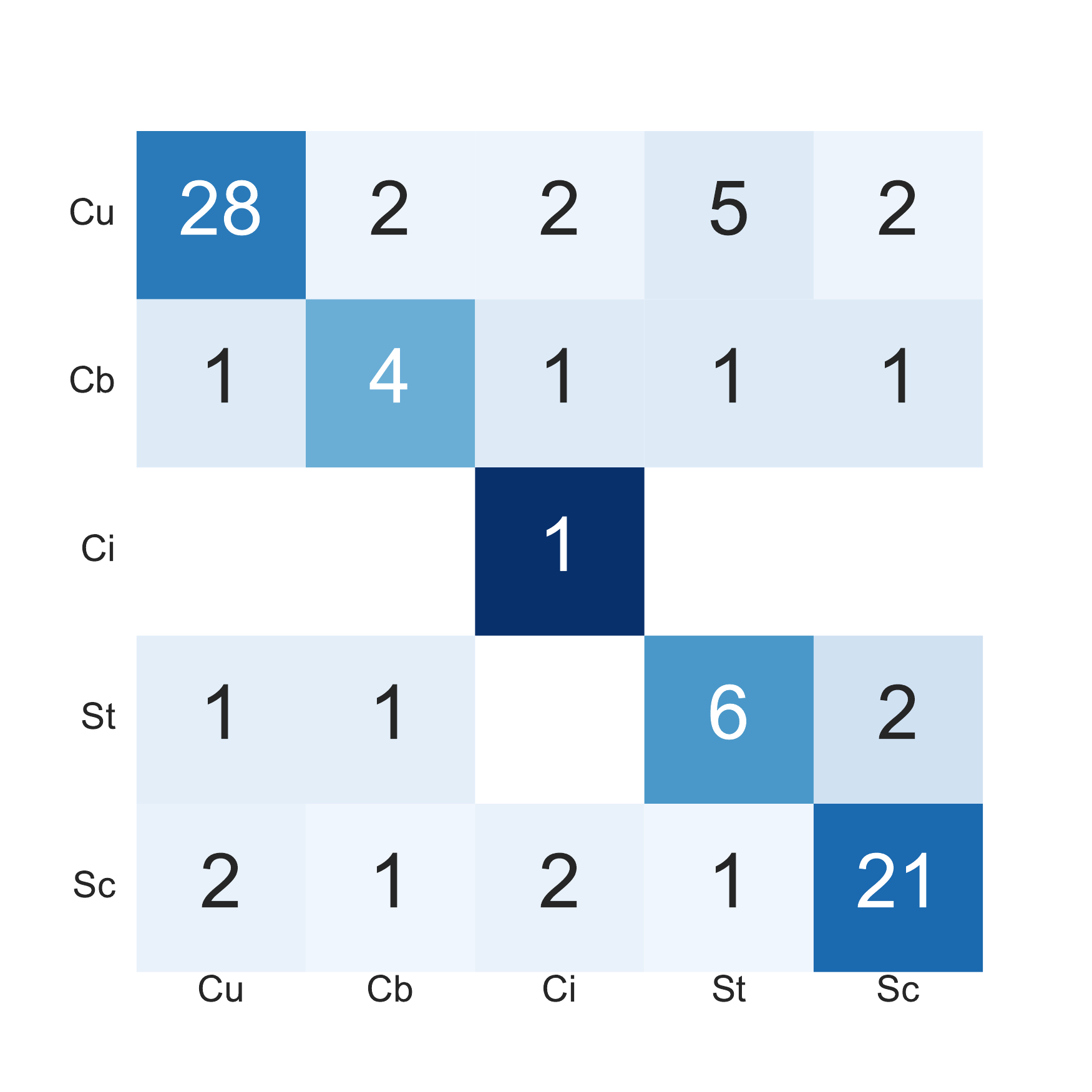} &
        \includegraphics[height=1in]{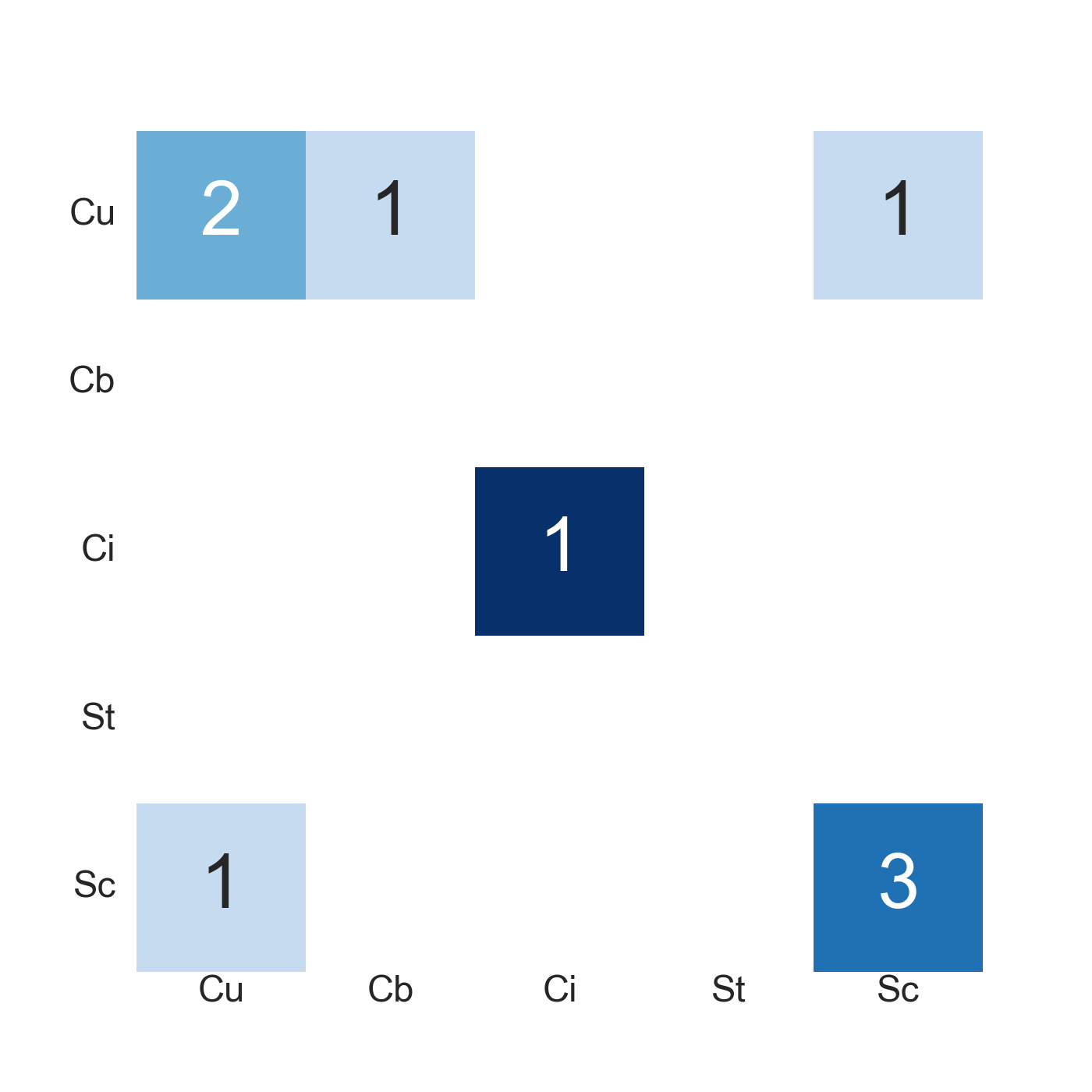} \\
        {J. Constable} & {Valenciennes} & {Lee} & {Watts}  & {L. Constable} & {Boudin} & {Cox} 
\end{tabular}\vspace{-0.1in}
\caption{The confusion matrices represent the classification results of all seven painters using the semi-supervised feature fusion model. The vertical axis represents the ground truth, while the horizontal axis represents the predicted labels. The abbreviations Cu, Cb, Cs, St, and Sc stand for cumuliform, cumulonimbiform, cirriform, stratiform, and stratocumuliform, respectively.}\label{fig:confusion_matrix}
\end{figure*}

Applying the trained SFF-CNN to the test photo images ($20\%$ of the CCSN dataset), we obtained a classification precision of $97.2\%$ and recall of $96.9\%$.
Detailed results on the test photos are provided in Section~\ref{photo}. Then, we re-trained the classification model on the entire CCSN dataset, which was then applied to the paintings. 
Because the painting dataset was small and the prevalence of different cloud types was highly unbalanced, to compute classification accuracy for the paintings, we only discriminated at the granularity of five common cloud types: cumuliform (cumulus), cumulonimbiform (cumulonimbus), cirriform (cirrus), stratiform (stratus, cirrostratus, altostratus, and nimbostratus), and stratocumuliform (cirrocumulus, altocumulus, and stratocumulus)\cite{barrett1976identification}. The classification accuracy of each painter using the SFF-CNN model with or without feature fusion is shown in Fig.~\ref{fig:bar_result}. For the accuracy achieved with feature fusion, the confidence interval for the accuracy at the significance level of $0.05$ is shown. Except for Cox, all the other artists had a confidence interval of accuracy well above $60\%$ (higher than the percentage of the most dominant cloud type), indicating that the clouds they painted correspond with Luke Howard's system of cloud categorization to a great extent. Moreover, clouds painted by Constable were the easiest to classify (highest accuracy) with a classification accuracy of 0.8452.
Additionally, in Fig.~\ref{fig:confusion_matrix}, we show the classification confusion matrices for each artist's paintings. Constable's clouds achieved the highest classification accuracy in the cumuliform.

To compare Constable with each of the other artists, we conducted hypothesis testing with the alternative hypothesis: Constable's paintings can be more accurately classified than those of other artists. We assigned identification numbers with Constable represented by $1$ and the other artists labeled as $2, 3, ..., 7$. We modeled the classification decision on a painting of the $i$th artist by a Bernoulli random variable with $1$ indicating the correct classification and $0$ otherwise. Let $p_i$ be the probability of correct classification. Thus, the distribution for the number of correctly classified paintings of artist $i$ is a Binomial distribution. The null hypothesis we formulated is $p_1\leq p_i$, $i\neq 1$. We used the one-tail Z-test~\cite{suissa1984uniformly} with continuity correction. The $p$-values obtained for Valenciennes, Lee, Watts, Lionel, Boudin, and Cox were $0.332$, $0.189$, $0.128$, $0.147$, $0.024$, and $0.189$, respectively. At the significance level of $0.1$, Constable's paintings were more accurately classified than Boudin's works, but not others. We conducted the same hypothesis testing to examine whether the inclusion of edge features could significantly improve the classification accuracy for any artist. The lowest $p$-value was $0.122$, obtained for Constable, while the other $p$-values exceeded $0.25$. This result indicates that the edge features improved classification most significantly for Constable.

From the results, it is evident that Constable's clouds correspond well with the system of cloud typology devised by Luke Howard. The $5\%$ confidence interval for the classification accuracy of Constable's paintings was $[0.768, 0.923]$. The average classification accuracy was highest for Constable's paintings. Are Constable's clouds more reminiscent of photographs of real-world clouds than those of his contemporaries? The answer is mixed. At the significance level of $0.1$, as indicated by the aforementioned $p$-values, Constable's clouds were more accurately classified than Boudin's, but not more than those by Valenciennes, Lee, Watts, Lionel Constable, and Cox. A potential explanation for the insignificant difference between Constable and these artists could be the limited number of paintings each of them had in the dataset.

We posit that Constable's technique, which involves strong contour lines rendered with a relatively continuous brushstroke, contributes to the realism of his clouds. In contrast, some artists, such as Boudin, tended to use dots and dashes in lieu of the clear-edged and smooth contours that define cloud shapes. Our computer model--trained on photographs--found Constable's cloud representations easier to classify and thus to recognize by viewers. Attention to precisely the morphological differences that Luke Howard highlighted when crafting his cloud typology in 1803 endowed Constable's clouds with a sufficiently striking degree of realism to set him apart from other landscape painters, at least in the eyes of his contemporaries--and in the eyes of our computer models. While our findings cannot confirm {\it definitively} that Constable was acquainted with Howard's cloud classification, they do confirm that systematic categorization is key for the visual impact of his realism.

\subsection{Style Similarity Analysis}
To train the style encoder for each artist, we used the MUNIT model~\cite{huang2018multimodal} as the network backbone. We excluded Learning Regression loss during training as suggested in~\cite{liu2020metrics} for better disentanglement of content and style features.
 All the paintings of an artist formed set ${A}$, and a subset of cloud photographs formed set ${P}$. We selected $300$ cloud photographs and ensured that the number of images in each cloud category was the same. For the paintings, instead of the original images, we used their sky-selected images. After obtaining the style encoders, we computed $D_{\text{style}}$ and $R_{\text{style}}$.
 
\subsubsection{Style Distance Between Artists' Clouds and Cloud Photos} We computed $R_\text{style}$ (defined earlier) for each of the seven painters. To assess variation in $R_\text{style}$ caused by randomness in the input images, for each painter, we randomly sampled five paintings to form a set and computed 
$R_\text{style}$ for this set. The calculation was repeated for multiple random samples of five paintings. As our collection only contained nine paintings by Cox, there were a maximum of $126$ different combinations of five paintings by Cox. We thus randomly sampled subsets of five paintings $126$ times for every artist. Table~\ref{tab:self} shows the average values of $R_\text{style}$ for each artist as well as the standard deviation.

To assess whether the distance metrics vary significantly among artists, we conducted hypothesis testing with the alternative hypothesis: these distances are significantly different between the artists. Denote the set of paintings from each of the seven artists by ${C}_i$, with $i=1,2,...,7$, and the sampled subsets by ${C}^n_i$, where $n=1,2, ..., 126$. Let the set ${R}_\text{style}({C}_i)=\{ R_\text{style}({C}_i^n): n\in\{1,2,...,126\}\}$.
Assume that the distribution of $R_\text{style}({C}_i^n)$ for each set ${C}_i$ follow a Gaussian distribution $N(\mu_i,\sigma^2_i)$, where $\mu_i$ and $\sigma^2_i$ indicate the mean and variance, respectively. Our null hypothesis is: $\mu_1=\mu_2,...,=\mu_7$. We use an $F$-test for a one-way analysis of variance. With an $F$-statistic of $21.15$ and a $p$-value below $2e-16$, the null hypothesis (the sets have the same mean value) is rejected at the significance level $0.05$. Then, we conducted another hypothesis test using the $T$-test to test if the paintings of Constable exhibit a style more akin to photographs compared with other artists. Let $\mu_1$ denote the mean value of Constable's painting set. We conducted six hypothesis tests with the null hypothesis: $\mu_1 \geq \mu_i$ for $i=2, 3, ..., 7$. Table~\ref{tab:self} shows both the $T$-statistics and the corresponding $p$-values. At a significance threshold of $0.1$, John Constable's painting style appears more similar to photographs than that of Boudin, Lee, and Cox. However, we cannot reject the null hypothesis that his painting style is less photo-like than that of Valenciennes, Lionel Constable, and Watt. In addition, we conducted the same $T$-test to determine whether, on average, $R_\text{style}$ of Valenciennes surpassed that of the other artists. All the $p$-values fell below 0.1. This result suggests that Valenciennes' painting style is the most reminiscent of actual photos when compared with the other six painters, at a significance level of 0.1. Furthermore, the Pearson correlation coefficient between classification accuracy and style similarity is -0.782, with a $p$-value of 0.039. This strong negative correlation between the measurement of stylistic difference (paintings versus photos) and the accuracy of cloud classification aligns well with our heuristic understanding--paintings similar to photos tend to be classified more accurately into cloud types. 

\begin{table}[ht!]
  \caption{ $R_\text{style}$ of the painting sets of each painter and $T$ statistics of $T$-test about the difference of $R_{\text{style}}$ between John Constable and other artists.}
    \vspace{-0.1in}
  \label{tab:self}
  \begin{tabular*}{\columnwidth}{@{\extracolsep{\fill}} lccc}
  \toprule
   Artist&$R_\text{style}$ ($\text{mean} \pm \text{std}$) & $T$-statistic & $p$-value\\
   \midrule
   Valenciennes& $1.163 \pm 0.132$ & 1.590 & 0.944\\
   Lionel Constable& $ 1.188 \pm 0.141$ & 0.165& 0.566\\
 John Constable & $1.191 \pm 0.147$ & -&-\\
 Watts& $1.210 \pm 0.146$ &-1.029 &0.152\\
Boudin & $1.254\pm 0.143$& -3.448&3.310e-04\\
Lee & $1.298 \pm  0.151$&-5.699&1.689e-08\\

Cox& $ 1.319 \pm 0.156$ &-6.703 &6.775e-11\\
\bottomrule
\end{tabular*}
\end{table}

\begin{table}[ht!]
  \caption{Style distance $D_{\text{style}}$ between painting set of John Constable with himself or others and $T$ statistics of $T$-test about the difference of $D_{\text{style}}$.}
  \vspace{-0.1in}
  \label{tab:pairs}
  \begin{tabular*}{\columnwidth}{@{\extracolsep{\fill}} lccc}
  \toprule
   Pair Comparison &$D_\text{style}$ ($\text{mean} \pm \text{std}$)&$T$-statistic & $p$-value\\
   \midrule
John Constable & $0.351 \pm 0.092$ & - & -\\
Lionel Constable&$0.359 \pm 0.095$ &-0.679 &0.249\\
Valenciennes&$0.373 \pm 0.109$ &-1.730 &0.042\\

Boudin&$ 0.405 \pm 0.108$&-4.270 &1.387e-05\\
Cox&$0.408 \pm 0.110$&-4.462 &6.225e-06\\
Watts&$0.421 \pm 0.113$&-5.390 &8.312e-08\\

Lee&$0.439 \pm 0.102$  &-7.190 &3.797e-12\\
 \bottomrule
\end{tabular*}
\end{table}

\subsubsection{Style Similarity Between Paintings by Constable and His Contemporaries}
Next, we used Eq.~(\ref{e3}) to compute the style similarity between pairs of painters. The results are shown in Table~\ref{tab:pairs}. Again, we conducted hypothesis testing to verify whether these style distances were significantly different. We used ${C}_1$ to denote the set of paintings by John Constable, and ${C}_i$ for those by another artist $i$. 
Similar to the approach in the previous subsection, we computed $D_\text{style}$ between randomly sampled subsets of paintings by two artists. The same subsets used to generate $R_\text{style}$ were used here. For the pair of sets ${C}_1$ and ${C}_i$, we obtained $126$ values of $D_\text{style}$: ${D}_\text{style}({C}_1, {C}_i)=\{D_\text{style}({C}_1^n, {C}_i^n), n\in \{1,2,...,126\}\}$. To establish a baseline, we also computed $D_{style}$ for subsets of paintings within John Constable's collection. Specifically, in addition to the $126$ subsets ${C}_1^n$ that were already created, another $126$ random subsets were sampled from ${C}_1$, each containing five paintings. Denote these new subsets by ${C}_{1,2nd}^n$, $n\in\{1,2,...,126\}$. Then, ${D}_\text{style}({C}_1, {C}_1)=\{ D_\text{style}({C}_1^n, {C}_{1,2nd}^n), n\in\{1,2,...,126\}\}$. If Constable's style significantly diverges from that of other artists in terms of ${D}_\text{style}$, we would expect the values in ${D}_\text{style}({C}_1, {C}_i)$ for $i\neq 1$ to surpass, at least on average, those in ${D}_\text{style}({C}_1, {C}_1)$. 

Denote the mean of ${D}_\text{style}({C}_1, {C}_i)$ by $\mu'_i$. In the first test, the null hypothesis is: $\mu'_1=\mu'_2,...,=\mu'_7$. Similarly, we used the $F$-test for one-way analysis of variance. The $F$-statistic obtained was $12.69$ with a $p$-value of $9.21e-14$, suggesting a significant difference in the style features among these paired artists.

The style distances between other artists and John Constable are provided in Table~\ref{tab:pairs}. We also conducted a T-test
between two data sets ${D}_\text{style}({C}_1, {C}_1)$ and ${D}_\text{style}({C}_1, {C}_i)$, where $i \in \{2,...11\}$ to test if artist $i$'s painting style is similar to John Constable's. The null hypothesis is: $\mu'_1\geq\mu'_i$ for $i\neq 1$. We tested at the confidence level of $0.95$. The $T$-statistic and the corresponding $p$-value for the $6$ tests are listed in Table~\ref{tab:pairs}, and we can observe that $p$-values are all below $0.05$ except for Lionel Constable. We can therefore claim that Lionel Constable's paintings are the most stylistically similar to John Constable's of the group. 

 
\subsection{Insights for Art History}
The key art-historical findings are:
(1) John Constable's clouds can be more accurately classified than those of his contemporaries, which sustains the possibility that Constable possessed some knowledge of Luke Howard's classification of clouds but does not serve as definitive proof.
(2) Fusing edge features boosts the classification performance of Constable's clouds more than it does for other artists. This underscores the significance of the pronounced structure in Constable's clouds as a contributing factor to their realistic portrayal.
(3) John Constable's paintings are not the most realistic among the artists evaluated if realism is defined by relative approximation in appearance to a photograph. Valenciennes, according to our experiments, created clouds that bear the closest resemblance to photographs.
(4) In terms of painting style, Lionel Constable aligns most closely with John Constable. This is consistent with his known practice of emulating his father's style.

\subsection{Classification Results on Cloud Photos}
\label{photo}



\begin{figure}[ht!]
  \centering
  \fbox{\includegraphics[trim={35 35 45 55},clip,width=0.4\linewidth]{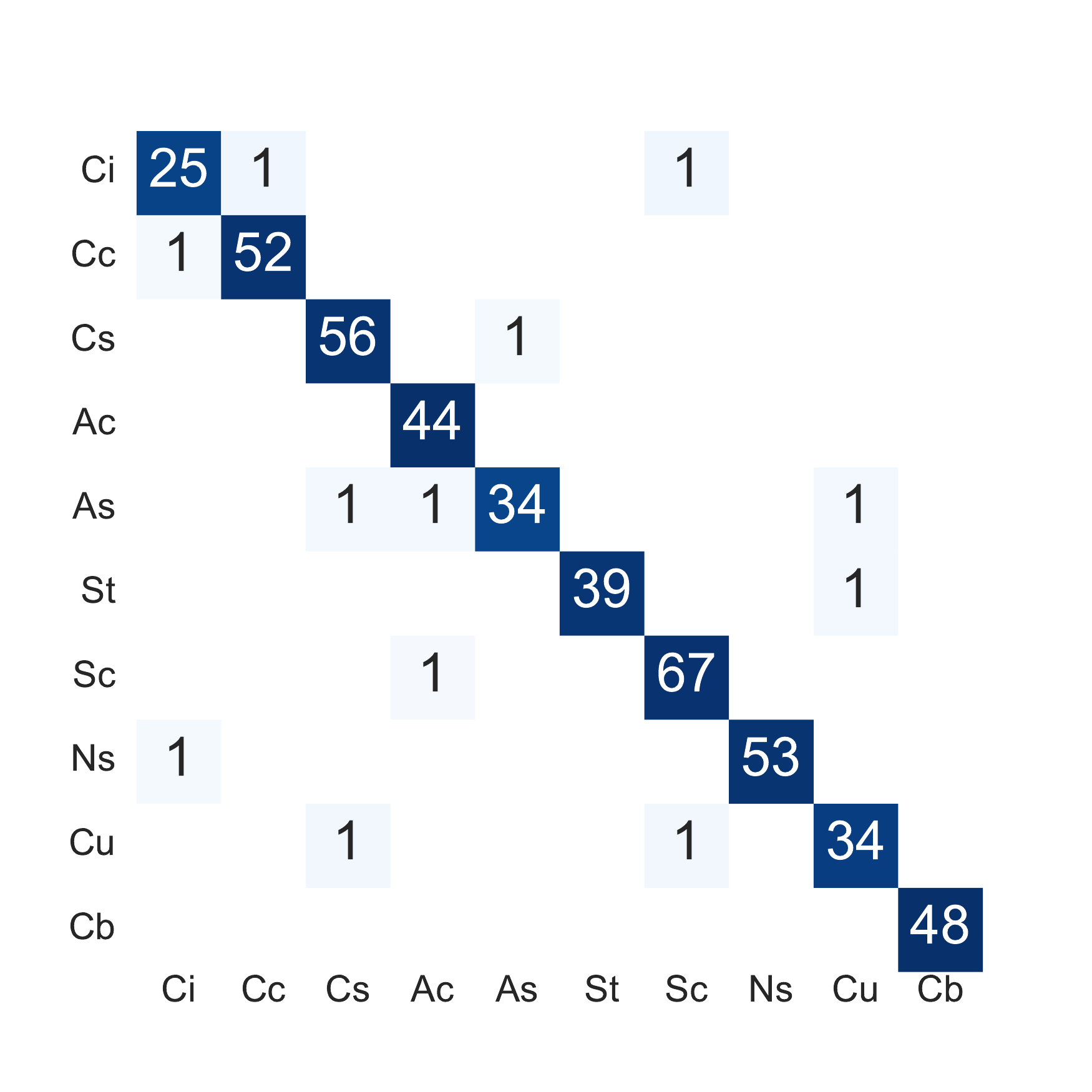}}\vspace{-0.1in}
  \caption{Confusion matrix of the test results on the CCSN dataset using our SFF-CNN model.}
  \label{fig:photo_cm}
\end{figure}

We randomly selected 20\% of the images from the CCSN dataset for testing. The other 80\% of the labeled images from the CCSN dataset and all the unlabeled images were used together during the self-learning process. In the training process, only parameters in the encoder for classic feature extraction were learned by back-propagation, while the parameters of the edge feature encoder were fixed. We chose Adam as the optimizer with a learning rate of $0.0001$ and batch size of $16$, which provided the highest accuracy. We compared the classification results obtained by our model with 
two advanced methods,  CloudNet~\cite{zhang2018cloudnet} and ensemble-learning-based classification~\cite{huertas2020cloud}. 
Our SFF-CNN model achieved the best performance with a precision of 0.972 and a recall of 0.969. The confusion matrix is shown in Fig.~\ref{fig:photo_cm}. In contrast, CloudNet (Ensemble learning) achieved a precision of 0.891 (0.953) and a recall of 0.868 (0.902). 
We also conducted the ablation study on the SFF-CNN model with results shown in Table~\ref{tab:ablation}. The improvement of the classification accuracy of SFF-CNN can be attributed to sky selection, the usage of unlabeled data, and edge feature fusion.

\begin{table}[ht]
  \caption{The ablation study of our model.}  \vspace{-0.1in}
  \label{tab:ablation}
  \begin{tabular*}{\columnwidth}{@{\extracolsep{\fill}} lcc}
  \toprule
   Method&Precision&Recall\\
   \midrule
    SFF-CNN ($\text{w}\slash \text{o}\ \text{feature fusion}$)&0.955&0.953\\
    SFF-CNN ($\text{w}\slash \text{o}\ \text{semi-supervised learning}$)&0.944&0.940\\
     SFF-CNN ($\text{w}\slash \text{o}\ \text{sky selection}$)&0.938&0.934\\
    SFF-CNN&\textbf{0.972}&\textbf{0.969}\\
    \bottomrule
\end{tabular*}
\end{table}

\section{Concluding Remarks}\label{sec:conclude}
Moving beyond investigating this artistic movement solely through traditional methods of art history or via computer-aided stylometric analysis, we engage with meteorology both as a means of gaining ground truth and as a historical discipline that may have influenced visual arts.

Following the assumption that the more realistic the cloud painting is, the easier it is for the AI to determine its cloud type, we developed a new, specialized computer-based cloud-type classification method to determine if Constable's clouds or those of his contemporaries can be correctly categorized into different cloud types. Additionally, by content-style disentanglement, we defined two metrics to evaluate the style similarity between paintings and photos as well as the similarity among artists.

Further avenues for art-historical inquiry are indicated by our research. The stylistic similarity between Valenciennes and Constable invites a reconsideration of their relationship. Our experiments suggest that even artists closely associated with naturalism like Boudin were working in a less photographic mode than like-minded predecessors who died just before photography was invented. This raises the interesting possibility that a kind of photographic realism was highly prized around 1800, but was soon seen as less realistic when applied to painting once photographs were more or less ubiquitous after the 1850s. These possibilities can be investigated further using the presented style similarity analysis. 

\ifCLASSOPTIONcompsoc
  \section*{Acknowledgments}
\else
  \section*{Acknowledgment}
\fi

This research was supported by the
National Endowment for the Humanities (NEH)
under Grant Nos. HAA-271801-20 and HAA-287938-22. 
This work used the Extreme Science and Engineering Discovery Environment (XSEDE), which was supported by the National Science Foundation (NSF) Grant No. ACI-1548562. J. Z. Wang has been supported by generous gifts from the Amazon Research Awards program and the NSF under Grant No. CNS-2234195. Xinye Zheng and Kevin R. Victor contributed in the early stage of the project. The authors would like to thank the Yale Center for British Art for hosting a visit of the team and for valuable discussions. They are grateful to the reviewers and editors for many constructive comments.


%

\ifCLASSOPTIONcaptionsoff
  \newpage
\fi



%



\bibliography{scibib}
\bibliographystyle{IEEEtran}

\ifCLASSOPTIONcaptionsoff
  \newpage
\fi

\begin{IEEEbiography}[{\includegraphics[width=0.9in,height=1.25in,clip,keepaspectratio,trim=0 8 0 1]{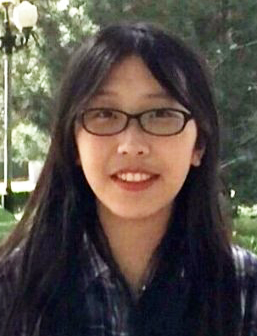}}]
{Zhuomin Zhang} is an Applied Scientist at Amazon in Seattle. She received her Ph.D. degree in Information Sciences and Technology from The Pennsylvania State University in 2022 under the joint supervision of Professors James Z. Wang and Jia Li. She received her bachelor's degree in Computer Science from Nanjing University in 2016. Her primary research interests include computer vision, machine learning, and multimedia. 
\end{IEEEbiography}

\begin{IEEEbiography}[{\includegraphics[width=0.9in,height=1.25in,clip,keepaspectratio,trim=90 0 110 0]{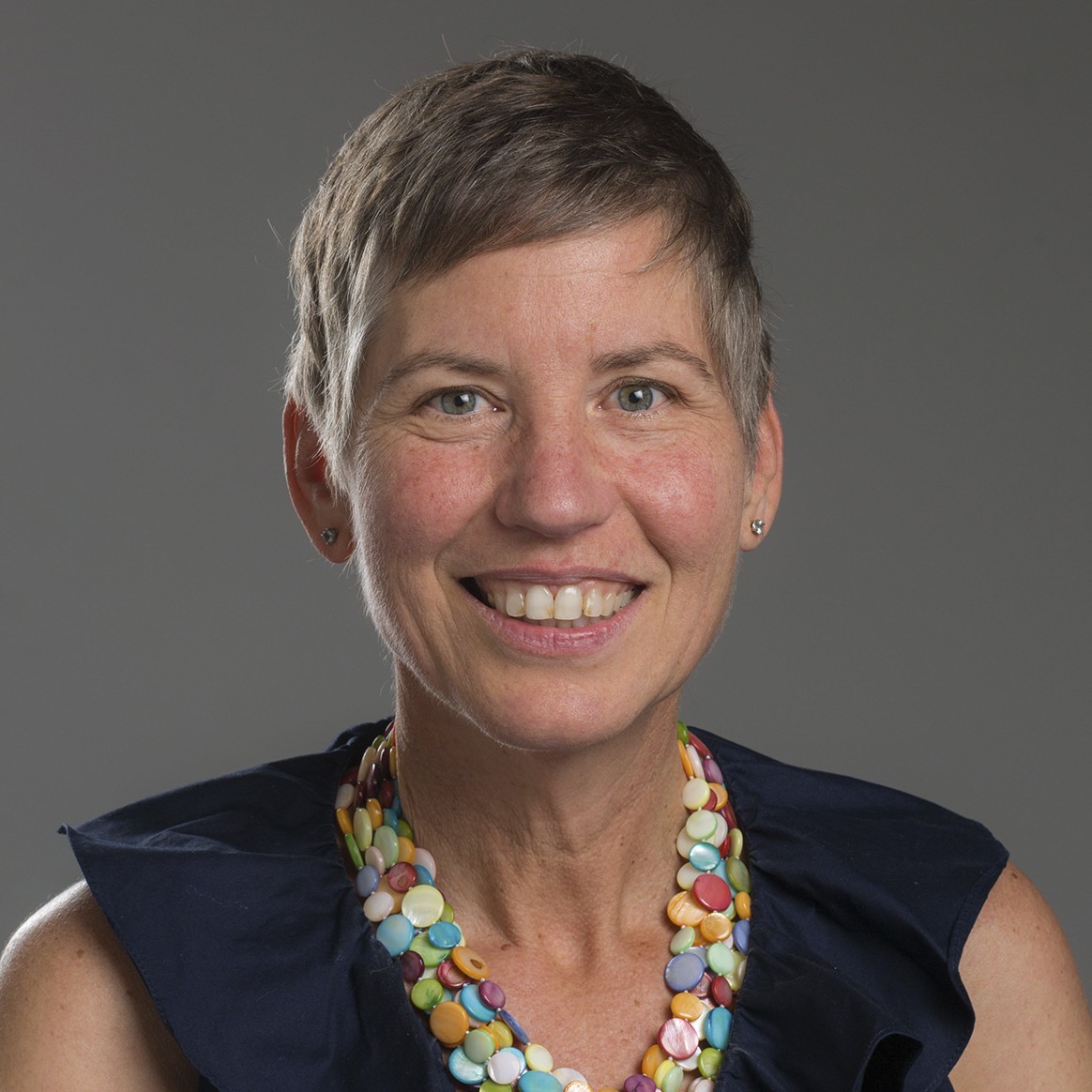}}]
{Elizabeth Mansfield} is Professor of Art History at Penn State. A specialist in 18th- and 19th-century European art and art historiography, her publications include the books The Perfect Foil: François-André Vincent and the Revolution in French Painting and Too Beautiful to Picture: Zeuxis, Myth, and Mimesis. The latter was awarded the Charles Rufus Morey book prize by the College Art Association in 2008. Past positions include academic appointments at New York University and at Sewanee: The University of the South. Prior to joining the faculty at Penn State in 2018, she served as Senior Program Officer at the Getty Foundation and Vice President for Scholarly Programs at the National Humanities Center. Her current book project is a study of Realism.  
\end{IEEEbiography}

\begin{IEEEbiography}[{\includegraphics[width=0.9in,height=1.25in,clip,keepaspectratio,trim=0 8 0 1]{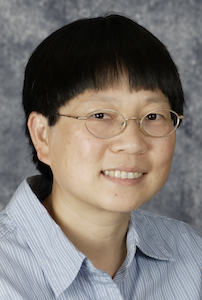}}] 
{Jia Li} (Fellow, IEEE) is a Professor of Statistics and (by courtesy) Computer Science at The Pennsylvania State University. Her research interests include machine learning, artificial intelligence, probabilistic graph models, and image analysis. She worked as a Program Director at the National Science Foundation from 2011 to 2013, a Visiting Scientist at Google Labs in Pittsburgh from 2007 to 2008, a researcher at the Xerox Palo Alto Research Center from 1999 to 2000, and a Research Associate in the Computer Science Department at Stanford University in 1999. She received the M.Sc. degree in Electrical Engineering (1995), the M.Sc. degree in Statistics (1998), and the Ph.D. degree in Electrical Engineering (1999), from Stanford University. She was Editor-in-Chief for Statistical Analysis and Data Mining: The ASA Data Science Journal from 2018 to 2020. She is a Fellow of the Institute of Electrical and Electronics Engineers and a Fellow of the American Statistical Association.
\end{IEEEbiography}

\begin{IEEEbiography}[{\includegraphics[width=0.9in,height=1.25in,clip,keepaspectratio,trim=0 0 0 0]{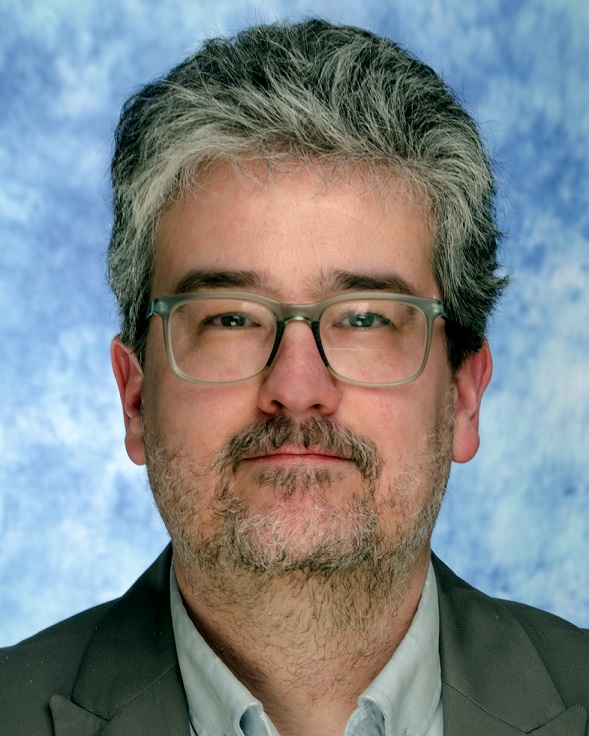}}]
{John Russell} is Associate Professor and Digital Humanities Librarian at The Pennsylvania State University Libraries and Associate Director of the Center for Virtual/Material Studies. He holds a B.A. from the University of Vermont and an MLS from Indiana University. 
\end{IEEEbiography}

\begin{IEEEbiography}[{\includegraphics[width=0.9in,height=1.25in,clip,keepaspectratio,trim=56 35 60 35]{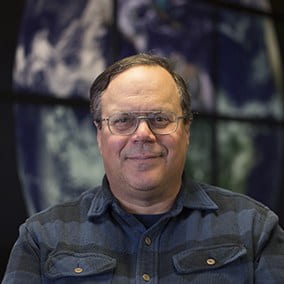}}]
{George Young} is a Professor of Meteorology at The Pennsylvania State University. His research interests are observational and statistical meteorology on scales ranging from individual clouds to large-scale weather systems. His contributions include participation in the planning, execution, and data analysis of numerous observational programs. His computational research involves the application of statistics, optimization, and machine learning to weather forecasting, satellite cloud image typing, and a range of interdisciplinary problems. He has been a member of the Penn State faculty since 1986. He received an MS degree in Meteorology (1982) from Florida State University and a PhD degree in Atmospheric Science (1986) from Colorado State University. He is a Fellow of the American Meteorological Society. 
\end{IEEEbiography}

\begin{IEEEbiography}[{\includegraphics[width=0.9in,height=1.25in,clip,keepaspectratio,trim=20 20 22 0]{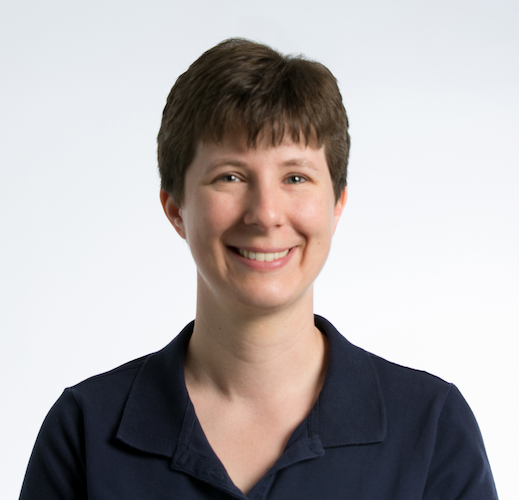}}]
{Catherine Adams} worked as the assistant curator of the Visual Resources Centre in the Department of Art History at The Pennsylvania State University from 2007 until 2021, when she became Digital Support Specialist in the new Center for Virtual/Material Studies. She has experience in metadata and digital image creation and curation. She has managed the Art History Department Visual Resource Collection and the Palmer Museum of Art's online collections in CONTENTdm. She aided in the creation of Arts and Architecture Resource Collaborative (AARC) and in several Omeka and Omeka S sites. She holds a B.A. from The Pennsylvania State University and an MLIS from the University of Pittsburgh. 
\end{IEEEbiography}

\begin{IEEEbiography}[{\includegraphics[width=0.9in,height=1.25in,clip,keepaspectratio,trim=0 20 0 0]{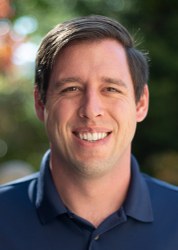}}]
{Kevin A. Bowley} received a B.S. degree in meteorology from Penn State University (2009) and a M.Sc. and Ph.D. degree in atmospheric and oceanic sciences from McGill University (2011 and 2017, respectively). He joined the faculty in the Department of Meteorology and Atmospheric Science at Penn State University in 2017 as a Lecturer, became an Assistant Teaching Professor in 2018, and is currently an Associate Teaching Professor (2023). His research and teaching interests center on understanding topics in synoptic meteorology, with a specific focus on Rossby wave breaking processes in the present and future climate. 
\end{IEEEbiography}
\vfill

\begin{IEEEbiography}[{\includegraphics[width=0.9in,height=1.25in,clip,keepaspectratio,trim=1 12 0 0]{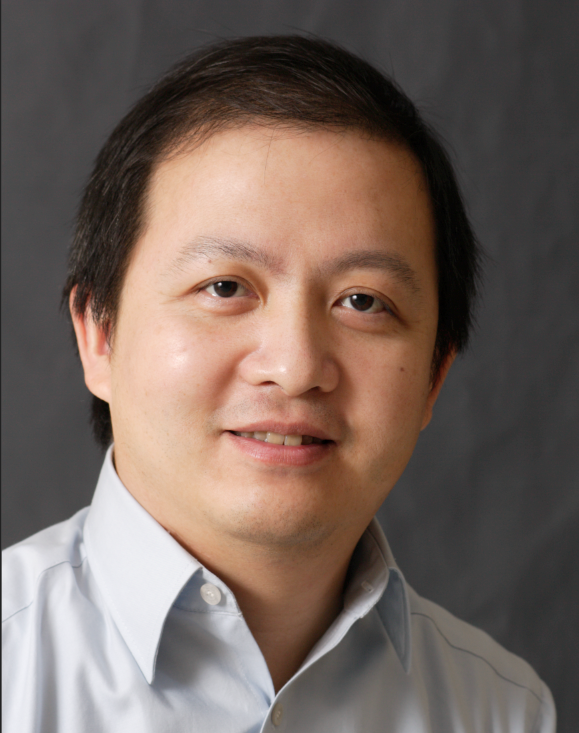}}]
{James Z. Wang} (Senior Member, IEEE)
is a Distinguished Professor of the Data Sciences and Artificial Intelligence section of the College of Information Sciences and Technology at The Pennsylvania State University. He received the bachelor's degree in mathematics {\it summa cum laude} from the University of Minnesota (1994), and the MS degree in mathematics (1997), the MS degree in computer science (1997), and the PhD degree in medical information sciences (2000), all from Stanford University. His research interests include image analysis, affective computing, image modeling, image retrieval, and their applications. He was a visiting professor at the Robotics Institute at Carnegie Mellon University (2007-2008), a lead special section guest editor of the IEEE Transactions on Pattern Analysis and Machine Intelligence (2008), a program manager at the Office of the Director of the National Science Foundation (2011-2012), and a special issue guest editor of the IEEE BITS -- The Information Theory Magazine (2022). He is also affiliated with the Department of Communication and Media, School of Social Sciences and Humanities, Loughborough University, UK (2023-2024). He was a recipient of a National Science Foundation Career Award (2004) and Amazon Research Awards (2018-2022).
\end{IEEEbiography}



\vfill


\end{document}